\documentclass[10pt,journal,cspaper,compsoc]{IEEEtran}

\usepackage[ruled,vlined]{algorithm2e}
\usepackage{amsmath}
\usepackage[T1]{fontenc}
\usepackage{yfonts}
\usepackage{cite}
\usepackage{xcolor}
\usepackage{graphicx}
\usepackage{epstopdf}
\usepackage{epsfig}
\usepackage{subfigure}
\usepackage{dsfont}
\usepackage{bm}
\usepackage{tikz}
\usetikzlibrary{fit}					
\usetikzlibrary{backgrounds}	
\usetikzlibrary{arrows}

\allowdisplaybreaks

\makeatletter

\newcommand{\Rmnum}[1]{\expandafter\@slowromancap\romannumeral #1@}
\makeatother

\begin{document}

\title{Dependent Indian Buffet Process-based Sparse Nonparametric Nonnegative Matrix Factorization}

\author{Junyu~Xuan,~Jie~Lu,~\IEEEmembership{Senior Member,~IEEE},~Guangquan~ Zhang,\\
Richard~Yi~Da~Xu,~Xiangfeng~Luo,~\IEEEmembership{Member,~IEEE}
\IEEEcompsocitemizethanks{
\IEEEcompsocthanksitem J. Xuan, J. Lu and G. Zhang are with the Centre for Quantum Computation and Intelligent Systems, School of Software, Faculty of Engineering
and Information Technology, University of Technology, Sydney (UTS), Australia and the School of Computer Engineering and Science, Shanghai University, China (e-mail: Junyu.Xuan@student.uts.edu.au; Jie.Lu@uts.edu.au; Guangquan.Zhang@uts.edu.au).
\IEEEcompsocthanksitem R. Y. D. Xu is with Faculty of Engineering
and Information Technology, University of Technology, Sydney (UTS). (e-mail: Yida.Xu@uts.edu.cn).
\IEEEcompsocthanksitem J. Xuan and X. Luo are with the School of Computer Engineering and Science, Shanghai University, China. (e-mail: luoxf@shu.edu.cn).
}
\thanks{}}

\IEEEcompsoctitleabstractindextext{%
\begin{abstract}
Nonnegative Matrix Factorization (NMF) aims to factorize a matrix into two optimized nonnegative matrices appropriate for the intended applications. The method has been widely used for unsupervised learning tasks, including recommender systems (rating matrix of users by items) and document clustering (weighting matrix of papers by keywords). However, traditional NMF methods typically assume the number of latent factors (i.e., dimensionality of the loading matrices) to be fixed. This assumption makes them inflexible for many applications. In this paper, we propose a nonparametric NMF framework to mitigate this issue by using dependent Indian Buffet Processes (dIBP). In a nutshell, we apply a correlation function for the generation of two stick weights associated with each pair of columns of loading matrices, while still maintaining their respective marginal distribution specified by IBP. As a consequence, the generation of two loading matrices will be column-wise (indirectly) correlated. Under this same framework, two classes of correlation function are proposed (1) using Bivariate beta distribution and (2) using Copula function. Both methods allow us to adopt our work for various applications by flexibly choosing an appropriate parameter settings. Compared with the other state-of-the art approaches in this area, such as using Gaussian Process (GP)-based dIBP, our work is seen to be much more flexible in terms of allowing the two corresponding binary matrix columns to have greater variations in their non-zero entries.
Our experiments on the real-world and synthetic datasets show that three proposed models perform well on the document clustering task comparing standard NMF without predefining the dimension for the factor matrices, and the Bivariate beta distribution-based and Copula-based models have better flexibility than the GP-based model.
\end{abstract}

}

\maketitle

\IEEEdisplaynotcompsoctitleabstractindextext

\IEEEpeerreviewmaketitle

\section{Introduction}

\IEEEPARstart{N}{onnegative} Matrix Factorization (NMF) is a renowned tool for factor analysis and unsupervised learning which has been used in many research areas \cite{5703094,6061964,6634167,6748996}. In text mining area, for example, factorization on the document-keyword matrix can discover the hidden \emph{topics} from documents, and these topics can assist document clustering or browsing; In the recommender system area, factorization on the user-movie matrix can help find the \emph{genres} of users and movies. Based on these \emph{genres}, the more accurate recommendation can be generated.

Sparse nonnegative matrix factorization \cite{ISI:000351063100027} is favoured by researchers in several areas due to its output: sparse representation of data. Sparse representation discovers a limited number of components to represent data, which is an important research problem \cite{heiler2006learning}. This sparse representation is generally desirable because it can aid human understanding (e.g., with gene expression data), reduce computational costs and obtain better generalization in learning algorithms. Sparsity is closely related to feature selection and certain generalizations in machine learning algorithms \cite{kim2008sparse,5593218}.

Although NMF has experienced a boom, with significant development in recent years, the assumption that number of dimensions of factor matrices need to be predefined is blocking its usage in real-world applications. Normally, this number (i.e., \emph{topic} number or \emph{genre} number) is assigned by experts with domain knowledge, but it seems that inaccurate assignment will impact on the performance of NMF on such applied tasks as document clustering or recommender systems. Furthermore, increase of the amount of data and complexity of the tasks means that even experts are inadequate for this job. Therefore, it is more reasonable and practical to automatically discover the topic number from the data.

Stochastic processes, such as the Dirichlet process \cite{neal2000markov} and Indian Buffet Process (IBP) \cite{griffiths2005infinite,griffiths2011indian}, have been used to replace ordinary distributions in nonparametric learning \cite{gershman2012tutorial} for providing priors of infinite vectors or matrices. In this context, IBP can be seen as the prior of a sparse matrix with an infinite number of columns. Therefore, an intuitive idea would be to give two factor matrices two IBPs priors, thus avoiding the need to predefine the dimensions for the factor matrices. However, the problem is how to ensure the two IBPs will generate the same dimension number for the two factor matrices if they are given to them separately.
A close related approach was seen to use GP-based dIBP \cite{griffiths2011indian}. This model works in the following fashion: We let $Z^{(1)}_j$ and $Z^{(2)}_j$ to be the jth corresponding columns of two identical sized binary matrices $Z^{(1)}$ and $Z^{(2)}$. The method assumes that both $Z^{(1)}_j$ and $Z^{(2)}_j$ share the same stick weight $\mu_j$, making the probability of generating the number of non-zero entries identical for both $Z^{(1)}_j$ and $Z^{(2)}_j$. The GP is merely used to control the correlations between individual entry pairs WITHIN $Z^{(1)}_j$ and $Z^{(2)}_j$, by thresholding a Gaussian CDF to make the entry to be either one or zero. Although this model improves upon the traditional NMF, it nonetheless is inadequate in many matrix factorisation scenarios as the assumption that the total number of non-zero entries are distributed identically does not hold universally. For example, the number of non-zero factors of a User column may be drastically different to the number of non-zero factors of a corresponding Item column in a recommender system.

In order to better and more flexibly describe the correlations between corresponding IBP columns, we instead propose a framework in which the correlation function is used for the generation of the latent weights pairs $\mu^{(1)}_j$ and $\mu^{(2)}_j$ which are responsible for the generation of $Z^{(1)}_j$ and $Z^{(2)}_j$ respectively. Compared with \cite{griffiths2011indian}, our work allow both $Z^{(1)}_j$ and $Z^{(2)}_j$ to have much greater flexibility and variations in terms of their non-zero entries.
To be specific, we propose two new dIBPs based on bi-variate beta distribution and copula. Instead of correlating two IBPs at the binary matrices level, the two proposed dIBPs correlate with the two IBPs at the very beginning (at the beta random variable level). This strategy results in the implementation of the nonparametric NMF framework being based on new dIBPs with a simple model structure. Another advantage of the new dIBPs is that the data correlation is directly modeled by the parameters of the bi-variate beta distribution and copula. Nevertheless, introducing bi-variate beta distribution and copula presents a challenge for the model inference. We have given three designed inference algorithms for three implementations of the nonparametric NMF framework: GP-based dIBP model, bi-variate beta distribution-based dIBP model, and copula-based dIBP model.
The experiments show that the proposed models perform well on the document clustering task without explicitly predefining the dimension number for the factor matrices, and the models based on new dIBPs have better convergence rates than GP-based dIBP.

The contributions of this paper are:

\begin{itemize}
  \item Two new dIBPs (i.e., bi-variate beta distribution-based and copula-based) with simpler model structure and direct correlation modeling capability are proposed as alternatives to the existing GP-based dIBP;
  \item Three dIBP-based nonparametric nonnegative matrix factorization models are proposed to remove the assumption of the traditional NMF that the number of factors needs to be fixed in advance.
\end{itemize}

The rest of this paper is organized as follows. Preliminary details of NMF and IBP are briefly introduced in Section 2. Section 3 reviews related work. Our dIBP-based nonparametric NMF framework is proposed with three implementations in Section 4, and Gibbs samplers are designed for the three models in Section 5. Section 6 conducts a set of experiments on real-world tasks. Lastly, Section 7 concludes the study and discusses further work.


\section{Preliminary Knowledge}

\subsection{Sparse Nonnegative Matrix Factorization}

Given a nonnegative matrix $Y_{m \times n}$ (extended to semi-nonnegative by \cite{Ding:2010}), Nonnegative Matrix Factorization (NMF) aims to find two matrices $A_{m \times k}$ and $X_{n \times k}$ to minimize the following cost function,
\begin{equation}
\begin{aligned}
J =& {\| Y_{m \times n} - A_{m \times k}X_{n \times k}^T \|}_F^2 + \|A_{m \times k}\|_1 + \|X_{n \times k}\|_1
\end{aligned}
\label{nmfcf}
\end{equation}
where $\| \cdot \|_F$ is the Frobenius norm, $\| \cdot \|_1$ is the $\ell_1$ norm and the elements of $A$ and $X$ are also nonnegative. The $\ell_1$ norm in the cost function is used for the sparseness constraint.

NMF is widely used in many different research areas. Take the recommender system as an example. The input $Y_{m \times n}$ denotes the ratings given by $m$ users on $n$ movies. The $A_{m \times k}$ denotes the users' interests, and $X_{n \times k}$ denotes the movies' genres. All users and movies are projected into the same $k$-dimensional space by NMF. Based on $A_{m \times k}$ and $A_{m \times k}$, a more accurate recommendation can be achieved.

One problem of NMF is to determine $k$. Normally, this variable is experimentally adjusted within a range. The problem of NMF without predefined $k$ is called Nonparametric Nonnegative Matrix Factorization.

\subsection{Indian Buffet Process}

The Indian Buffet Process (IBP) \cite{griffiths2005infinite,griffiths2011indian} is defined as a prior for the binary matrices with an infinite number of columns. The graphical model is shown in Fig. \ref{fig:gm}(a). A stick-breaking construction for IBP \cite{teh2007stick} is proposed as follows:
\begin{equation}
\begin{aligned}
\nu_j \sim Beta(\alpha, 1),~
\mu_k = \prod_{j=1}^{k} \nu_j,~
z_{n,k} \sim& Bernoulli(\mu_k)
\end{aligned}
\label{sbibp}
\end{equation}
where ${z_{n,k}}$ form a matrix $Z_{N \times K}$, $\nu_j$ is the variable with a beta distribution, and $\alpha$ is the parameter of beta distribution. $\mu_k$ is the stick weight of column $k$. The bigger $\mu_k$ is, the more `ones' there are in the column $k$ of the binary matrix $Z_{N \times K}$. The final number of $K$ is determined by the data and the parameter $\alpha$ of IBP.

An intuitive idea of applying IBP for Nonparametric Nonnegative Matrix Factorization is to separately set two IBPs as priors of $A$ and $X$. One problem with this idea is that the number of columns $K_1$ and $K_2$ generated by the two IBPs may be different, which violates the requirement of NMF.


\section{Related Work}

\begin{table}[!t]%
\centering
\caption{Notations in this paper}{%
\begin{tabular}{c|p{6.5cm}}
\hline
Symbol & meaning in this paper
\\\hline
$M$         & the row number of $Y$ \\\hline
$N$         & the column number of $Y$ \\\hline
$K$         & the latent factor number \\\hline
$Y$         & data matrix with size $M \times N$\\\hline
$y_{m,n}$   & the element of $Y$ at $m$ row and $n$ column \\\hline
$A$         & factor matrix with size $M \times K$ \\\hline
$X$         & factor matrix with size $N \times K$ \\\hline
$a_{m,k}$   & the element of $A$ at $m$ row and $k$ column \\\hline
$x_{n,k}$   & the element of $X$ at $n$ row and $k$ column \\\hline
$Z^{(1)}$       & mask matrix for $A$ \\\hline
$Z^{(2)}$       & mask matrix for $X$ \\\hline
$z^{(1)}_{m,k}$   & the mask binary variable for element of $A$ at $m$ row and $k$ column \\\hline
$z^{(2)}_{n,k}$   & the mask binary variable for element of $X$ at $n$ row and $k$ column \\\hline
$V^{(1)}$         & loading matrix for $A$ \\\hline
$V^{(2)}$         & loading matrix for $X$ \\\hline
$v^{(1)}_{m,k}$   & the loading variable for element of $A$ at $m$ row and $k$ column \\\hline
$v^{(2)}_{n,k}$   & the loading variable for element of $X$ at $n$ row and $k$ column \\\hline
$\nu^{(1)}_{k}$   & $k$ stick weight of IBP for $A$  \\\hline
$\nu^{(2)}_{k}$   & $k$ stick weight of IBP for $X$ \\\hline
$\theta$   & model parameters for Bivariate beta distribution or copula\\\hline
\end{tabular}}
\label{tb:notations}
\end{table}%

The related work mainly falls into two categories: one concerns research on the Nonnegative Matrix Factorization and the other concerns about the Indian Buffet Process. These state-of-the-art researches inspire our idea of using a dependent Indian Buffet Process for nonparametric nonnegative matrix factorization.

\subsection{Nonparametric Nonnegative Matrix Factorization}

Types of research on the Nonnegative Matrix Factorization include supervised or semi-supervised extension\cite{Ding:2010,huh2013supervised}, the convergence rate\cite{4359171}, Sparse\cite{4801604}, Nonparametric\cite{liang2013beta} and more.

There are also some researches on Bayesian extension of NMF which aim to model the NMF using the distributions. In one example, LDA \cite{blei2003latent} and correlated LDA \cite{blei2007correlated} ideas are transferred to the Bayesian parametric NMF and correlated NMF \cite{paisleybayesian}. These extensions are still parametric, however, which means the dimension number still need to be predefined.

The nonparametric extension of NMF mainly relies on the machinery of stochastic processes. The Beta process is used as the prior of one factor matrix in \cite{liang2013beta}. The Gamma process is used to generate the coefficients for the combination of corresponding elements in factor matrices rather than the prior of the factor matrices \cite{blei2010bayesian}. Both have been successfully applied for music analysis. IBP is used as the prior of one factor matrix and another factor matrix is drawn from Gaussian distribution; an efficient inference method (Power-EP) is proposed for this model\cite{ding2010nonparametric}. These processes can be considered as the extension of the latent feature factor model \cite{griffiths2011indian}. However, there is no work to place the priors for two factor matrices simultaneously.

\subsection{IBP and dIBP}

The idea of dependent nonparametric processes was first proposed by MacEachern \cite{maceachern2000dependent}. Seven classes of dependence are summarized \cite{foti2013survey}. The basic IBP is proposed in \cite{griffiths2005infinite,griffiths2011indian}, and is actually a marginalization of the Beta-Bernoulli process. Its widespread popularity is due to its power to generate a binary matrix with infinite columns. The dIBP was first proposed in \cite{williamson2010dependent} based on Gaussian process, and can be used for the nonparametric NMF after being embedded in our proposed framework which is discussed in Section 4. The Kernel Beta process \cite{ren2011kernel} is another dependent beta process model that models the dependence between data points with different covariants, such as the time tags of documents, geographic locations of people or GDPs of countries. Hierarchial beta process is proposed so that the different beta processes share a common base discrete measure \cite{thibaux2007hierarchical}. The Phylogenetic Indian Buffet Process \cite{miller2012phylogenetic} considers the tree structure of the data points, which can be seen as a supervised IBP. A coupled IBP \cite{chatzis2012coupled} is proposed for collaborative filtering, which links two IBPs through a factor matrix. Therefore, this coupling does not guarantee that the factor matrices have the same dimension number which is important for the NMF. dimension number of two infinite factor matrices .  However, there is no work on using or constructing dIBP for nonparametric NMF.


\begin{figure*}
\subfigure[Indian Buffet Process]{
\begin{minipage}[b]{\textwidth}
    \tikzstyle{rv}=[circle,
                        thick,
                        minimum size=1.2cm,
                        draw=black!80,
                        ]
    \tikzstyle{line}=[->,
                        solid,
                        line width=1pt,
                        draw=black!80,
                        ]

    \begin{tikzpicture}[font=\Large,scale=0.9,rotate=90]
     	

    	\node[rv] (1) at (0, 0) {$\alpha$};

    	\draw[thick, rounded corners] (-1.3,-7.5) rectangle (1.3,-1);

    	\node[rv] (3) at (0, -3) {$\mu_k$};
        \draw[thick, rounded corners] (-1,-7) rectangle (1,-5);
    	\node[rv] (4) at (0, -6) {$z_{n, k}$};

    	\path[line]
    		(1)		edge (3)
    		(3)		edge (4);

    \end{tikzpicture}
\end{minipage}
}
\subfigure[Implementation by Gaussian Processes-based dependent Indian Buffet Process]{
\begin{minipage}[b]{\textwidth}
    \tikzstyle{rv}=[circle,
                        thick,
                        minimum size=1.2cm,
                        draw=black!80,
                        ]
    \tikzstyle{line}=[->,
                        solid,
                        line width=1pt,
                        draw=black!80,
                        ]
    \tikzstyle{dt}=[circle,
                        thick,
                        minimum size=1.2cm,
                        draw=black!80,
                        fill=gray
                        ]
    \begin{tikzpicture}[font=\Large,scale=0.9,rotate=90]
     	

    	\node[rv] (alpha) at (0, 0) {$\alpha$};

    	\draw[thick, rounded corners] (-3.3,-11.5) rectangle (3.3,-1);

    	\node[rv] (muk) at (0, -3) {$\mu_k$};

        \draw[thick, rounded corners] (-3.2,-3.3) rectangle (-0.8,-11);
        \node[rv] (gamank1) at (-2.2, -4.5) {$\Gamma^{(1)}_{n^{(1)}, k}$};
    	\node[rv] (hnk1) at (-2.2, -7) {$h^{(1)}_{n^{(1)}, k}$};
        \node[rv] (znk1) at (-1.8, -10) {$z^{(1)}_{n^{(1)}, k}$};

        \draw[thick, rounded corners] (3.2,-3.3) rectangle (0.8,-11);
        \node[rv] (gamank2) at (2.2, -4.5) {$\Gamma^{(2)}_{n^{(2)}, k}$};
    	\node[rv] (hnk2) at (2.2, -7) {$h^{(2)}_{n^{(2)}, k}$};
        \node[rv] (znk2) at (1.8, -10) {$z^{(2)}_{n^{(2)}, k}$};

        \node[rv] (gk)  at (0, -8) {$g_k$};
        \node[rv] (sigmak) at (0, -10) {$\Sigma_k$};

        \node[rv] (v)  at (0, -12.5) {$\tau$};

        \node[rv] (V1)  at (-2.2, -13.5) {$V^{(1)}$};
        \node[rv] (V2) at (2.2, -13.5) {$V^{(2)}$};

        \node[rv] (a)  at (-1, -15.5) {$X$};
        \node[rv] (x) at (1, -15.5) {$A$};

        \node[dt] (y) at (0, -17.5) {$Y$};

    	\path[line]
    		(alpha)		edge (muk)

    		(muk)		edge (znk1)
            (muk)		edge (znk2)
            (hnk1)		edge (znk1)
            (hnk2)		edge (znk2)
            (gk)	    edge (hnk1)
            (gk)	    edge (hnk2)
            (sigmak)    edge (gk)
            (gamank1)   edge (hnk1)
            (gamank2)   edge (hnk2)
            (v)         edge (V1)
                        edge (V2)
            (V1)        edge (a)
            (V2)        edge (x)
            (znk1)      edge (a)
            (znk2)      edge (x)
            (a)         edge (y)
            (x)         edge (y)
        ;

    \end{tikzpicture}
\end{minipage}
}
\subfigure[Implementations by Bivariate-based or Copula-based dependent Indian Buffet Process]{
\begin{minipage}[b]{\textwidth}
    \tikzstyle{rv}=[circle,
                        thick,
                        minimum size=1.2cm,
                        draw=black!80,
                        ]
    \tikzstyle{line}=[->,
                        solid,
                        line width=1pt,
                        draw=black!80,
                        ]
    \tikzstyle{dt}=[circle,
                        thick,
                        minimum size=1.2cm,
                        draw=black!80,
                        fill=gray
                        ]
    \begin{tikzpicture}[font=\Large, scale=0.9, rotate=90]
     	

    	\node[rv] (1) at (0, 0) {$\mathbf{\theta}$};

    	\draw[thick, rounded corners] (-2.3,-7.5) rectangle (2.3,-1) ;
    	\node[rv] (2) at (-1.2, -3) {$\mu_k^{(1)}$};
        \node[rv] (7) at (1.2, -3) {$\mu_k^{(2)}$};

        \draw[thick, rounded corners] (-0.2,-5) rectangle (-2.2,-7);
    	\node[rv] (z1) at (-1.2, -6) {$z_{n^1, k}^{(1)}$};

        \draw[thick, rounded corners] (0.2,-5) rectangle (2.2,-7);
        \node[rv] (z2) at (1.2, -6) {$z_{n^2, k}^{(2)}$};

        \node[rv] (v) at (0, -9) {$\tau$};

        \node[rv] (V1) at (-1.8, -11) {$V^{(1)}$};
        \node[rv] (V2) at (1.8, -11) {$V^{(2)}$};

        \node[rv] (X) at (-0.8, -13) {$X$};
        \node[rv] (A) at (0.8, -13) {$A$};

        \node [dt] (y) at (0, -15) {$Y$};
    	\path[line]
    		(1)		edge (2)
            		edge (7)
    		(2)		edge (z1)
            (7)		edge (z2)
            (z1)		edge (X)
            (z2)		edge (A)
            (V1)		edge (X)
            (V2)		edge (A)
            (v)		edge (V1)
            		edge (V2)
            (A)		edge (y)
            (X)		edge (y)
            ;
    \end{tikzpicture}
\end{minipage}
}
 \caption{Graphical Models for (a) the original Indian Buffet Process, (b) GP-based dependent Indian Buffet Process and (c) Bivariate beta distribution-based or Copula-based dependent Indian Buffet Process}
 \label{fig:gm}
\end{figure*}
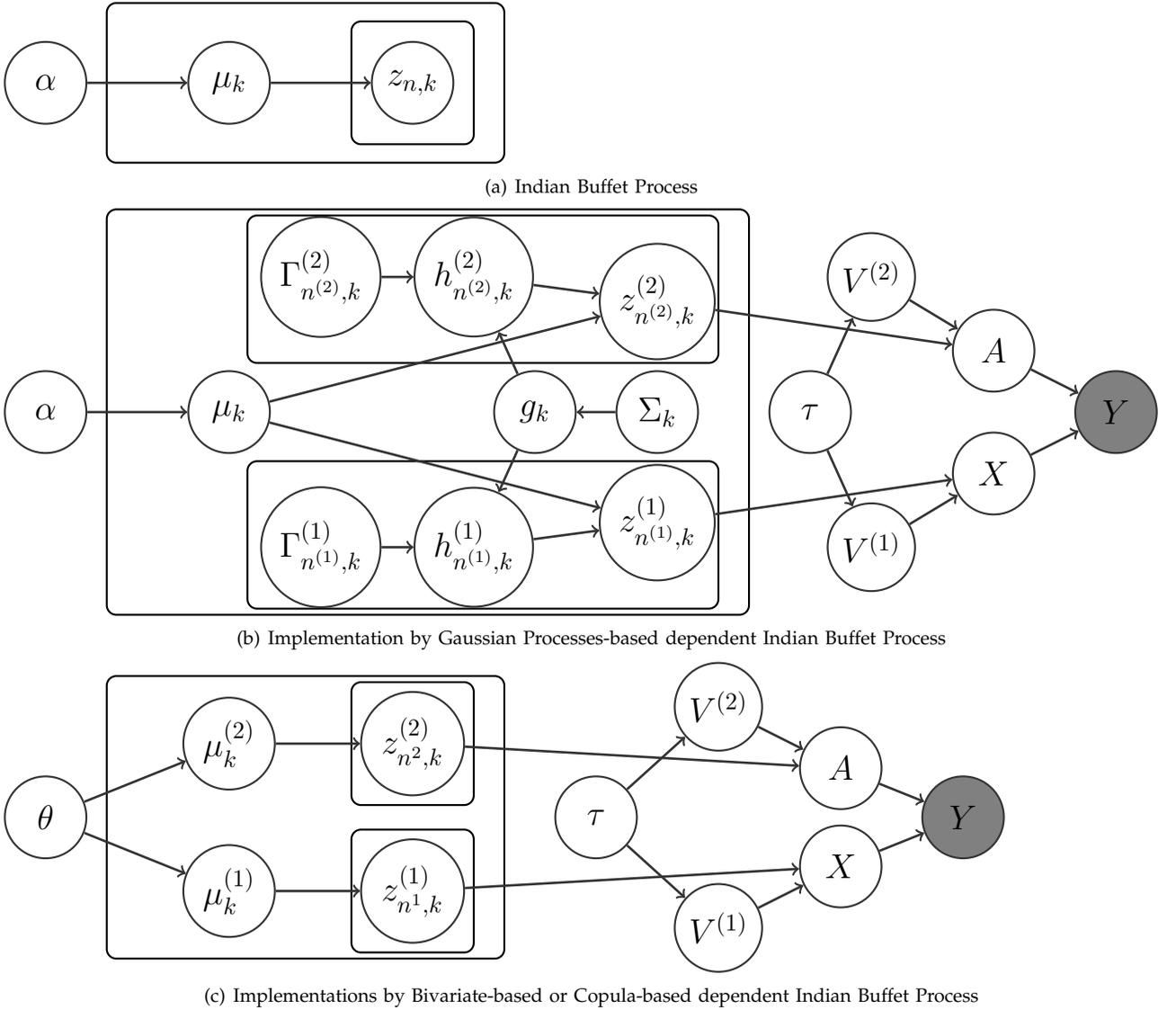

\section{Nonparametric NMF Framework and three implementations}

To set a prior for the infinite matrices $A$ and $X$ in NMF and make sure the two matrices have the same number of columns, we apply the dependent Indian Buffet process as the prior for two matrices.

Nonparametric NMF Framework, the data matrix is modeled as,
\begin{equation}
\begin{aligned}
Y = A * X^T = (V^{(1)} \odot Z^{(1)})*(V^{(2)} \odot Z^{(2)})^T
\end{aligned}
\label{nmfmodel}
\end{equation}
where $Z^{(1)}$ and $Z^{(2)}$ are two binary matrices, $V^{(1)}$ and $V^{(2)}$ are loading matrices, and $\odot$ denotes the Hadamard product. A dIBP is used as the prior for $Z^{(1)}$ and $Z^{(2)}$.
The likelihood of nonnegative matrix factorization is,
\begin{equation}
\begin{split}
a_{m, k} = v^{(1)}_{m, k} \cdot z^{(1)}_{m, k}&, ~v^{(1)}_{m, k} \sim gamma(1, \tau_1)
\\
x_{n, k} = v^{(2)}_{n, k} \cdot z^{(2)}_{n, k}&, ~v^{(2)}_{n, k} \sim gamma(1, \tau_2)
\\
y_{m, n} | a_{m, k}, x_{n, k} \sim& Exp(y_{m, n}; \sum_k a_{m, k} \cdot x_{n, k} + \epsilon)
\end{split}
\label{nmflikelihood}
\end{equation}
where $Exp()$ denotes the exponential distribution, the selection of the gamma and exponential distributions for $V$ is used to guarantee the nonnegativity of $A$ and $X$, and each element of $Y$ satisfies an exponential distribution with support $[0, +\infty)$. The special parameters for the distributions are to retain the desired expectations of these distributions. For example, the expectation of distribution $y_{m, n}$ is $\sum_k a_{m, k} \cdot x_{n, k} + \epsilon$.
The matrices $V^{(1)}$ and $V^{(2)}$ are two loading matrices. Since IBP can only generate two binary matrices $Z^{(1)}$ and $Z^{(2)}$, we need $V^{(1)}$ and $V^{(2)}$ to approximate the data $Y$. Finally, the $(V^{(1)} \odot Z^{(1)})$ and $(V^{(2)} \odot Z^{(2)})$ can be seen as the $A$ and $X$ in Eq. (\ref{nmfmodel}). The number of columns is not predefined but is learned from the data.

The implementation of nonnegative matrix factorization is only conducted to select a prior for $Z^{(1)}$ and $Z^{(2)}$. In the following subsections, we will introduce three implementations of this framework through three dIBPs.

\subsection{Implementation by GP-based dIBP}

The first dIBP is proposed based on Gaussian Process (GP) \cite{williamson2010dependent}. In this GP-based dIBP model, each stick weight $\mu_k$ is used to generated a different number of columns of binary matrices. For the NMF, we can use this model as the prior for the matrices $A$ and $X$. The graphical model is shown in Fig. \ref{fig:gm}(b), and the generative process is as follows,
\begin{equation}
\begin{aligned}
\nu_j &\sim Beta(\alpha, 1),~~
\mu_k &= \prod_{j=1}^{k} \nu_j
\end{aligned}
\label{gpibp1}
\end{equation}
where the $\mu_k$ are IBP sticks as in Eq. (\ref{sbibp}). This set of sticks is shared by two binary matrices through
\begin{align}
g_k &\sim GP(0, \Sigma_k)\nonumber
\\\nonumber
h^{(1)}_{n_1,k} &\sim GP(g_k, \Gamma^{(1)}_{n_1,k})
\\\nonumber
h^{(2)}_{n_2,k} &\sim GP(g_k, \Gamma^{(2)}_{n_2,k})
\\\nonumber
\Sigma_k(t, t^\prime) &= \sigma^2 \exp(-\frac{(t - t^\prime)^2}{s^2})
\\\nonumber
\Gamma^{(1)} &= \Gamma^{(2)} = \eta^2 I
\\\nonumber
z^{(1)}_{n_{(1)},k} = \delta\{ h^{(1)}_{n_1,k} &< F^{-1}(\mu_k | 0, (\Sigma_k)^{(1, 1)} + (\Gamma^{(1)}_{n_1,k})^{(1,1)}) \}
\\\nonumber
z^{(2)}_{n_{(2)},k} = \delta\{ h^{(2)}_{n_2,k} &< F^{-1}(\mu_k | 0, (\Sigma_k)^{(2, 2)} + (\Gamma^{(2)}_{n_2,k})^{(2,2)}) \}
\end{align}
where $F(\cdot)$ is the normal cumulative distribution function, $\delta\{\cdot\}$ is the indicator function, and $\Sigma_k$ is the kernel function for the GP. The model details can be found in \cite{williamson2010dependent}. Since there are only two binary matrices, the GP is degenerated to a two-dimensional Gaussian distribution, and $\Sigma_k$ is a $2 \times 2$ matrix.

\subsection{Implementation by Bivariate Beta Distribution-based dIBP}

There are different ways to link two IBPs. The GP-based IBP uses the same set of sticks for different binary matrices. Here, we propose another method that links the initials of two IBPs, $\nu$. In the original IBP, $\nu$ satisfies a beta distribution with parameter $(\alpha, 1)$. Therefore, we use a joint distribution $(\nu^{(1)}, \nu^{(2)})$ with beta distributions $Beta(\alpha_1, 1)$ and $Beta(\alpha_2, 1)$ as marginal distributions. Following this idea, the intuitive candidate would be Dirichlet distribution. However, there is a strictly negative relation, $\nu^1 + \nu^2 = 1$, between the samples of the Dirichlet distribution, but we hope to preserve the freedom of two $(\nu^{(1)}, \nu^{(2)})$.

Instead of the Dirichlet distribution, a
bivariate beta distribution \cite{Olkin2003407} is adopted. The probability density function is defined as,
\begin{equation}
\begin{aligned}
p(x, y) &= \frac{x^{a-1} y^{b-1} (1-x)^{b+c-1} (1-y)^{a+c-1}}
{B(a,b,c) (1-xy)^{a+b+c}}
\\
&s.t., ~~0 \le x, y \le 1, ~~ a, b, c > 0
\end{aligned}
\label{bibeta}
\end{equation}
where $a, b, c$ are three parameters of this distribution. One merit of this bivariate beta distribution is that two marginal distributions are,
\begin{equation}
\begin{aligned}
x &\sim Beta(a, c),~~
y &\sim Beta(b, c)
\end{aligned}
\label{bibetamd}
\end{equation}
Another metric is that this distribution models positive correlation between $(\mu^{(1)}, \mu^{(2)})$ with range $[0, 1]$ adjusted by the three parameters $(a, b, c)$ compared to the Dirichlet distribution. Here, we set $c$ of the bivariate beta distribution to 1. The reason is because we must ensure that the marginal distribution of each $\nu$ is a beta distribution with parameter $(\alpha, 1)$. This condition is for the distribution of binary matrices generated by the model that satisfies the IBPs. Considering Eq. (\ref{bibetamd}), we give $c$ a fixed value. Even with a fixed value for $c$, the bivariate beta distribution in Eq. (\ref{bibeta}) is still able to model different correlations of two variables. For example, when $a=2.5$ and $b=4$, the correlation between $x$ and $y$ is $0.978$; when $a=0.05$ and $b=0.1$, the correlation between $x$ and $y$ is $0.080$.

With the desired bivariate distribution in hand, we build the model as follows,
\begin{equation}
\begin{split}
(\nu_k^{(1)}, \nu_k^{(2)}) ~\sim~ biBeta(\mathbf{\theta})&, ~~\mathbf{\theta} : \{ a, b, c=1\} > 0
\\
z_{n,k}^{(1)} ~\stackrel{\text{i.i.d}}{\sim}~ Bernoulli(\mu_k^{(1)})&,~~ \mu_k^{(1)} = \prod_{j=1}^k \nu_j^{(1)}
\\
z_{n,k}^{(2)} ~\stackrel{\text{i.i.d}}{\sim}~ Bernoulli(\mu_k^{(2)})&,~~ \mu_k^{(2)} = \prod_{j=1}^k \nu_j^{(2)}
\end{split}
\label{model}
\end{equation}
where $biBeta(\cdot)$ denotes the bivariate beta distribution in Eq. (\ref{bibeta}) and $\theta$ denotes the parameters of the distribution. The graphical model is shown in Fig. \ref{fig:gm}(c).

The likelihood function is the same as Eq. (\ref{nmflikelihood}). Although the bivariate beta distribution-based dIBP has extended the freedom of the relation between $\nu^{(1)}$ and $\nu^{(2)}$, the relation is restricted to positive relations in bivariate beta distribution. We use the copula to capture more freedom of relations.

\subsection{Implementation by Copula-based dIBP}

\begin{figure*}[!th]
  \centerline{\includegraphics[scale=0.5]{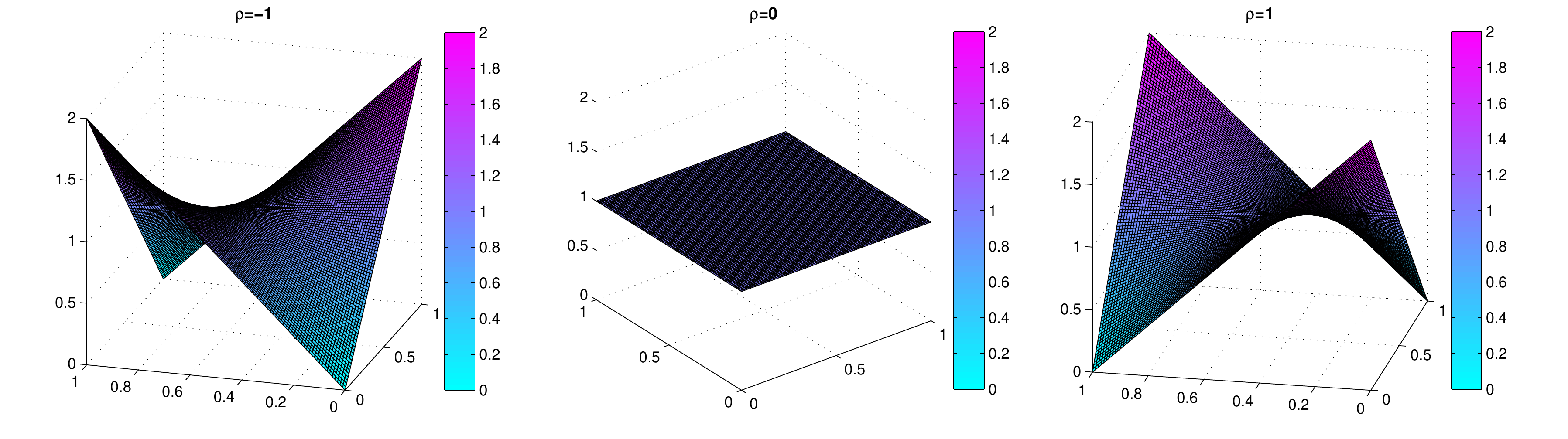}}
  \caption{FGM copula density surface with different values of $\rho = \{-1, 0, 1\}$ and the marginal distributions are both beta distributions: $beta(1, 1)$ (means that $\alpha_1 = \alpha_2 = 1$ in Eq. (\ref{fgmcopula})). }
  \label{fig:fgmcopula}
\end{figure*}

Copula \cite{trivedi2007copula} is another alternative for linking two variables with given marginal distributions, and is used to define a joint distribution for variables with known marginal distributions. Here, we select the Farlie-Gumbel-Morgenstern (FGM) Copula \cite{beare2010copulas} as an example.

The definition of the FGM Copula is,
\begin{equation}
\begin{split}
    C_{\rho}(u, v) &= uv + \rho uv(1-u)(1-v),~~ \rho \in [-1, 1] \\
    c_{\rho}(u, v) &= 1 + \rho (2u-1)(2v-1) \\
\end{split}
\label{fgmcopula}
\end{equation}
where
\begin{equation}
\begin{aligned}
    u = F(\nu^{(1)}) &\sim (\nu^{(1)})^{\alpha_1},~& f(\nu^{(1)}) \sim \alpha_1 \cdot (\nu^{(1)})^{\alpha_1 - 1}\\
    v = F(\nu^{(2)}) &\sim (\nu^{(2)})^{\alpha_2},~&
    f(\nu^{(2)}) \sim \alpha_2 \cdot (\nu^{(2)})^{\alpha_2 - 1}
\end{aligned}
\label{uv}
\end{equation}
and $C_{\rho}(u, v)$ is the cumulative distribution function, $c_{\rho}(u, v)$ is the probability density function, and $\rho$ is the parameter of the FGM copula. $u$ and $v$ are two marginal distributions that are known in advance. For our dIBP model, these marginal distributions are beta distribution with parameters $(\alpha_1, 1)$ and $(\alpha_2, 1)$.

The correlation or dependence of $\nu^{(1)}$ and $\nu^{(2)}$ is modeled or reflected by the value of $\rho$. As shown in Fig. \ref{fig:fgmcopula}, the different correlation (i.e., positive or negative) can be captured by the value of $\rho$. In particular, if $\rho=0$, there is no correlation between two marginal distributions.

FGM-based dIBP is defined by replacing the bivariate beta distribution in Eq. (\ref{model}) with joint distribution defined by the FGM copula,
\begin{equation}
\begin{split}
    p(\nu_k^{(1)}, \nu_k^{(2)}) ~&=~ c_{\rho}(\mathbf{\theta}),
    \\
    ~~&\mathbf{\theta} : \{ \rho \in [-1, 1], \alpha_1 > 0, \alpha_2 > 0 \} \\
\end{split}
\end{equation}
The graphical model is the same as the bivariate beta distribution in Fig. \ref{fig:gm}(c). The Spearman correlation can be evaluated by $coco = \frac{\rho}{3}$. Since the support of $\rho$ is $[-1, 1]$, the correlation range that can be modelled by the FGM Copula is $[-\frac{1}{3}, \frac{1}{3}]$.

Comparing the bivariate beta distribution, the advantage of copulas is that we can easily obtain the cumulative distribution function and probability density function of $(\nu_k^{(1)}, \nu_k^{(2)})$, because we only know the exact form of probability density function of the bivariate beta distribution but not the exact form of the cumulative distribution function. This will impact on the model inference, which will be introduced later.

One advantage of bivariate beta distribution-based and copula-based dIBP compared with GP-based dIBP that less latent variables are involved. This is easily observed through the graphical models in Fig. \ref{fig:gm}. More latent variables will slow the convergence of the Gibbs sampler.


\section{Model Inference}

With data $Y$ in hand, the objective of this section is to estimate the hidden variables by a properly designed MCMC sampler for their posterior distribution, $p(\bm{\mu}, \mathbf{Z}, \mathbf{V}, \bm{\theta} | Y)$. It is difficult to perform posterior inference under infinite mixtures, thus a common work-around solution in nonparametric Bayesian learning is to use a truncation method. The truncation method, which uses a relatively big $K$ as the (potential) maximum number of factors, is widely accepted.

Note that GP-based dIBP is not our contribution. Our contribution is to link the GP-based dIBP with the nonnegative matrix factorization likelihood. The inference of this model, which is not our contribution, is given in the Appendix for the self-contained purpose. The details can be found in \cite{williamson2010dependent}.

\subsection{Update $\bm{\mu}$}

When updating $\mu_k$, we need to find the conditional distribution of $\mu_k$ given $\mu_{k-1}$ and $\mu_{k+1}$, because the order of $\mu$ must be maintained to make the marginal distributions of $Z$ satisfy two IBPs. The acceptance ratio of the M-H sampler for $\mu_{(k)}=[\mu_{(k)}^{(1)}, \mu_{(k)}^{(2)}]$ is,
\begin{equation}
\begin{split}
\text{min} \left (1,
\frac{
p(Z|\mu^*_{(k)})\cdot p(\mu^*_{(k)}|\mu_{(k+1)}, \mu_{(k-1)}) }
{p(Z|\mu_{(k)}) \cdot p(\mu_{(k)}|\mu_{(k+1)}, \mu_{(k-1)}) }
\times \frac{q(\mu_{(k)})}{q(\mu^*_{(k)})}
\right )
\end{split}
\label{mhmu}
\end{equation}
where $p(Z|\mu_{(k)})$ is the likelihood of $\mu_{(k)}$ to generate $k$th column of binary matrix $Z$.

The $p(\mu^*_{(k)}|\mu_{(k+1)}, \mu_{(k-1)})$ in Eq. (\ref{mhmu}) is the conditional probability density function of $\mu^*_{(k)}$ within the range of $[\mu_{(k+1)}, \mu_{(k-1)}]$. This will be different when different strategies are used to link $\mu^{(1)}_{(k)}$ and $\mu^{(2)}_{(k)}$.
Next, we derive the conditional probability density function of $\mu_{(k)}$ as follows:

For the first column,
\begin{equation}
\begin{aligned}
\mu^{(1)}_1 = \nu^{(1)}_1,~~\mu^{(1)}_2 = \nu^{(1)}_2
\end{aligned}
\end{equation}
and
\begin{equation}
\begin{aligned}
p(\mu^{(1)}_1, \mu^{(1)}_2)=p(\nu^{(1)}_1, \nu^{(1)}_2)
\end{aligned}
\end{equation}

For the second column,
\begin{equation}
\begin{aligned}
\mu^{(2)}_1 = \nu^{(2)}_1 \mu^{(1)}_1,~~
\mu^{(2)}_2 = \nu^{(2)}_2 \mu^{(1)}_2
\end{aligned}
\end{equation}
and
\begin{equation}
\begin{aligned}
p(\mu^{(2)}_1, \mu^{(2)}_2) &= p(\nu^{(2)}_1\cdot \mu^{(1)}_1, \nu^{(2)}_2\cdot \mu^{(1)}_2)
\\
&= \frac{p(\nu^{(2)}_1, \nu^{(2)}_2) }{ |J^{(1)}| }
\\
&= \frac{p(\frac{\mu^{(2)}_1}{\mu^{(1)}_1}, \frac{\mu^{(2)}_2}{\mu^{(1)}_2})}{ \mu^{(1)}_1 \cdot \mu^{(1)}_2}
\end{aligned}
\end{equation}
where $J^{(1)}$ is the Jacobian matrix,
\begin{equation}
J^{(1)} =
\begin{bmatrix}
       \mu^{(1)}_1 & 0           \\[0.3em]
       0 & \mu^{(1)}_2
\end{bmatrix}
\end{equation}

For the $n$th column,
\begin{equation}
\begin{aligned}
\mu^{(n)}_1 = \nu^{(n)}_1 \mu^{(n-1)}_1,~~\mu^{(n)}_2 = \nu^{(n)}_2 \mu^{(n-1)}_2
\end{aligned}
\end{equation}
and
\begin{equation}
\begin{aligned}
p(\mu^{(n)}_1, \mu^{(n)}_2)&=p(\nu^{(n)}_1\cdot \mu^{(n-1)}_1, \nu^{(n)}_2\cdot \mu^{(n-1)}_2)
\\
&= \frac{p(\nu^{(n)}_1, \nu^{(n)}_2) }{ |J^{(n-1)}| }
\\
&= \frac{p(\frac{\mu^{(n)}_1}{\mu^{(n-1)}_1}, \frac{\mu^{(n)}_2}{\mu^{(n-1)}_2})}{ \mu^{(n-1)}_1 \cdot \mu^{(n-1)}_2}
\end{aligned}
\end{equation}
where $J^{(n-1)}$ is the Jacobian matrix,
\begin{equation}
J^{(n-1)} =
\begin{bmatrix}
       \mu^{(n-1)}_1 & 0           \\[0.3em]
       0 & \mu^{(n-1)}_2
\end{bmatrix}
\end{equation}
To summarize, the conditional density of $\mu^{(n)}$ is
\begin{equation}
\begin{aligned}
&p(\mu^{(n)} | \mu^{(n-1)}, \mu^{(n+1)})
\\
=& \frac{p(\frac{\mu^{(n)}_1}{\mu^{(n-1)}_1}, \frac{\mu^{(n)}_2}{\mu^{(n-1)}_2})}{ \mu^{(n-1)}_1 \cdot \mu^{(n-1)}_2}
\cdot
\frac{p(\frac{\mu^{(n+1)}_1}{\mu^{(n)}_1}, \frac{\mu^{(n+1)}_2}{\mu^{(n)}_2})}{ \mu^{(n)}_1 \cdot \mu^{(n)}_2}
\end{aligned}
\end{equation}

The different methods of linking $\nu_1$ with $\nu_2$ will lead to a different joint probability density function for them. As introduced in Section 4, we have given two specific forms of their joint probability in Eq. (\ref{bibeta}) and Eq. (\ref{fgmcopula}).

The proposal distribution, $q(\cdot)$, is selected as the product of two independent truncated beta distributions,
\begin{equation}
\begin{split}
q(\mu_{k}) &= beta(\mu^{(1)}_{k}; \frac{\alpha_1}{K}, 1)\cdot beta(\mu^{(2)}_{k}; \frac{\alpha_2}{K}, 1) \\
\mu^{(1)}_{k} &\sim beta(\frac{\alpha_1}{K}, 1),   ~ \mu^{(1)}_{k} \in [\mu^{(1)}_{k+1}, \mu^{(1)}_{k-1}]\\
\mu^{(2)}_{k} &\sim beta(\frac{\alpha_2}{K}, 1),   ~ \mu^{(2)}_{k} \in [\mu^{(2)}_{k+1}, \mu^{(2)}_{k-1}]
\end{split}
\end{equation}

\subsection{Update $\mathbf{Z}$}

Two binary matrices, $\mathbf{Z}:\{Z^{(1)}, Z^{(2)}\}$, can be updated separately, since each element in two matrices satisfies a Bernoulli distribution with the following conditional probabilities,
\begin{equation}
\begin{aligned}
&~~~p(z^{(1)}_{m, k} = 1)
\\
&\propto \mu^{(1)}_{k}  \prod_n Exp(y_{m, n}; \sum_l a_{m, l} \cdot x_{n, l} + \epsilon)
\\
&\propto \mu^{(1)}_{k}  \prod_n Exp(y_{m, n}; \sum_{l} v^{(1)}_{m, l}  z^{(1)}_{m, l}  v^{(2)}_{n, l}  z^{(2)}_{n, l} + \epsilon)
\\
&~~~p(z^{(1)}_{m, k} = 0)
\\
&\propto (1 - \mu^{(1)}_{k})  \prod_n Exp(y_{m, n}; \sum_l a_{m, l} \cdot x_{n, l} + \epsilon)
\\
&\propto (1 - \mu^{(1)}_{k})  \prod_n Exp(y_{m, n}; \sum_l v^{(1)}_{m, l} z^{(1)}_{m, l} v^{(2)}_{n, l} z^{(2)}_{n, l} + \epsilon)
\end{aligned}
\label{bbnmfz1}
\end{equation}
and
\begin{equation}
\begin{aligned}
&~~~p(z^{(2)}_{n, k} = 1)
\\
&\propto \mu^{(2)}_{k} \prod_m Exp(y_{m, n}; \sum_l a_{m, l} \cdot x_{n, l} + \epsilon)
\\
&\propto \mu^{(2)}_{k}  \prod_m Exp(y_{m, n}; \sum_{l} v^{(1)}_{m, l}  z^{(1)}_{m, l}  v^{(2)}_{n, l}  z^{(2)}_{n, l} + \epsilon)
\\
&~~~p(z^{(2)}_{n, k} = 0)
\\
&\propto (1 - \mu^{(2)}_{k}) \prod_m Exp(y_{m, n}; \sum_l a_{m, l} \cdot x_{n, l} + \epsilon)
\\
&\propto (1 - \mu^{(2)}_{k})  \prod_m Exp(y_{m, n}; \sum_{l} v^{(1)}_{m, l}  z^{(1)}_{m, l}  v^{(2)}_{n, l}  z^{(2)}_{n, l} + \epsilon)
\end{aligned}
\label{bbnmfz2}
\end{equation}
where $\epsilon$ is small positive number.

\subsection{Update $\mathbf{V}$}

Since the prior for $V$ is Gamma distribution and the likelihood is exponential distribution, the conditional distribution for $\mathbf{V}$ is,
\begin{equation}
\begin{aligned}
&~~~~p(v^{(1)}_{m, k} | z^{(1)}_{m, k}, a_{m, k})
\\
&\propto gam(v^{(1)}_{m, k} ; \alpha_1) \prod_n Exp(y_{m, n}; \sum_k a_{m, k} \cdot x_{n, k} + \epsilon)
\\
&\propto gam(v^{(1)}_{m, k} ; \alpha_1) \prod_n Exp(y_{m, n}; \sum_k v^{(1)}_{m, k} z^{(1)}_{m, k} v^{(2)}_{n, k} z^{(2)}_{n, k} + \epsilon)
\end{aligned}
\label{nmflikelihoodv1}
\end{equation}
and
\begin{equation}
\begin{aligned}
&~~~~p(v^{(2)}_{n, k} | z^{(2)}_{n, k}, x_{n, k})
\\
&\propto gam(v^{(2)}_{n, k} ; \alpha_2) \prod_m Exp(y_{m, n}; \sum_l a_{m, l} \cdot x_{n, l} + \epsilon)
\\
&\propto gam(v^{(2)}_{n, k} ; \alpha_2) \prod_m Exp(y_{m, n}; \sum_l v^{(1)}_{m, l} z^{(1)}_{m, l} v^{(2)}_{n, l} z^{(2)}_{n, l} + \epsilon)
\end{aligned}
\label{nmflikelihoodv2}
\end{equation}

\subsection{Update $\bm{\theta}$}

In Fig. \ref{fig:gm}(c), the graphical model has a parameter $\theta$. For the different strategies to link $(\nu^{(1)}, \nu^{(2)})$, the parameters must be different. Therefore, we design corresponding update methods for the proposed two strategies: bivariate beta distribution and copula.

\subsubsection{Bivariate Beta Distribution}

The parameters of bivariate beta distribution, $\bm{\theta}:\{a, b\}$, are given two gamma priors. The conditional distributions are,
\begin{equation}
p([a~ b]| \cdots) \propto gam([a~ b];hp) \prod_{k=1}^K p(\mu^{(1)}_k, \mu^{(2)}_k| a, b)
\label{bbnmfab}
\end{equation}
where $hp$ is the hyper-parameter of the prior for $a$ and $b$.

\subsubsection{Copula}

There are three parameters for each copula, $\bm{\theta}:\{\rho, \alpha_1, \alpha_2\}$. Their conditional distributions are,
\begin{equation}
p([\alpha_1~ \alpha_2]| \cdots) \propto gam([\alpha_1~ \alpha_2];hp) \prod_{k=1}^K c(\mu^{(1)}_k, \mu^{(2)}_k)
\label{cnmfalpha}
\end{equation}

We give the $\rho$ of the FGM copula a uniform distribution on $[-1, 1]$,
\begin{equation}
p(\rho | \cdots) \propto \prod_{k=1}^K c(\mu^{(1)}_k, \mu^{(2)}_k | \rho)
\label{cnmfrho}
\end{equation}

After introducing the update methods, we summarize the inference for the three models in Algorithm \ref{ag:gpnmf} for the GP-based dIBP NMF (GP-dIBP-NMF) model, Algorithm \ref{ag:bbdnmf} for the Bivariate beta distribution-based dIBP NMF (BB-dIBP-NMF) model, and Algorithm \ref{ag:cpnmf} for the Copula-based dIBP NMF (C-dIBP-NMF) model.

\begin{algorithm}[!t]
\caption{Gibbs Sampler for GP-dIBP-NMF}
\KwIn{$Y$}
\KwOut{$A$, $X$}
initialization\;
\While{$i \le max_{iter}$}
{
    // \textit{latent variables of dIBP}\\
    Update $\mu$ by Eq. (\ref{gpnmfmu});\\
    Update $Z$ by Eq. (\ref{gpnmfz1}) and (\ref{gpnmfz2});\\
    Update $g$ by Eq. (\ref{gpnmfg});\\
    Update $h$ by Eq. (\ref{gpnmfh});\\
    Update $s$ by Eq. (\ref{gpnmfs});\\

    // \textit{latent variables of data of NMF}\\
    Update $V$ by Eq. (\ref{nmflikelihoodv1}) and (\ref{nmflikelihoodv2});\\

    $i++$;
}

Select the sample with largest likelihood;\\
return $A$ and $X$;
\label{ag:gpnmf}
\end{algorithm}

\begin{algorithm}[!t]
\caption{Gibbs Sampler for BB-dIBP-NMF}
\KwIn{$Y$}
\KwOut{$A$, $X$}
initialization\;
\While{$i \le max_{iter}$}
{
    // \textit{latent variables of dIBP}\\
    Update $\mu$ by Eq. (\ref{mhmu});\\
    Update $Z$ by Eq. (\ref{bbnmfz1}) and (\ref{bbnmfz2});\\
    Update $a$ and $b$ by Eq. (\ref{bbnmfab});\\

    // \textit{latent variables of data of NMF}\\
    Update $V$ by Eq. (\ref{nmflikelihoodv1}) and (\ref{nmflikelihoodv2});\\

    $i++$;
}

Select the sample with largest likelihood;\\
return $A$ and $X$;
\label{ag:bbdnmf}
\end{algorithm}

\begin{algorithm}[!t]
\caption{Gibbs Sampler for C-dIBP-NMF}
\KwIn{$Y$}
\KwOut{$A$, $X$}
initialization\;
\While{$i \le max_{iter}$}
{
    // \textit{latent variables of dIBP}\\
    Update $\mu$ by Eq. (\ref{mhmu});\\
    Update $Z$ by Eq. (\ref{bbnmfz1}) and (\ref{bbnmfz2});\\
    Update $\alpha_1$ and $\alpha_2$ by Eq. (\ref{cnmfalpha});\\
    Update $\rho$ by Eq. (\ref{cnmfrho});\\

    // \textit{latent variables of data of NMF}\\
    Update $V$ by Eq. (\ref{nmflikelihoodv1}) and (\ref{nmflikelihoodv2});\\

    $i++$;
}

Select the sample with largest likelihood;\\
return $A$ and $X$;
\label{ag:cpnmf}
\end{algorithm}


\section{Experiments}

In this section, we first express the ability to conduct matrix factorization and hidden factor number learning of the proposed algorithms through synthetic examples, followed by the performance of the algorithms on real-world tasks.

\subsection{Synthetic dataset}

\begin{figure}[t]
  \centerline{\includegraphics[scale=0.6]{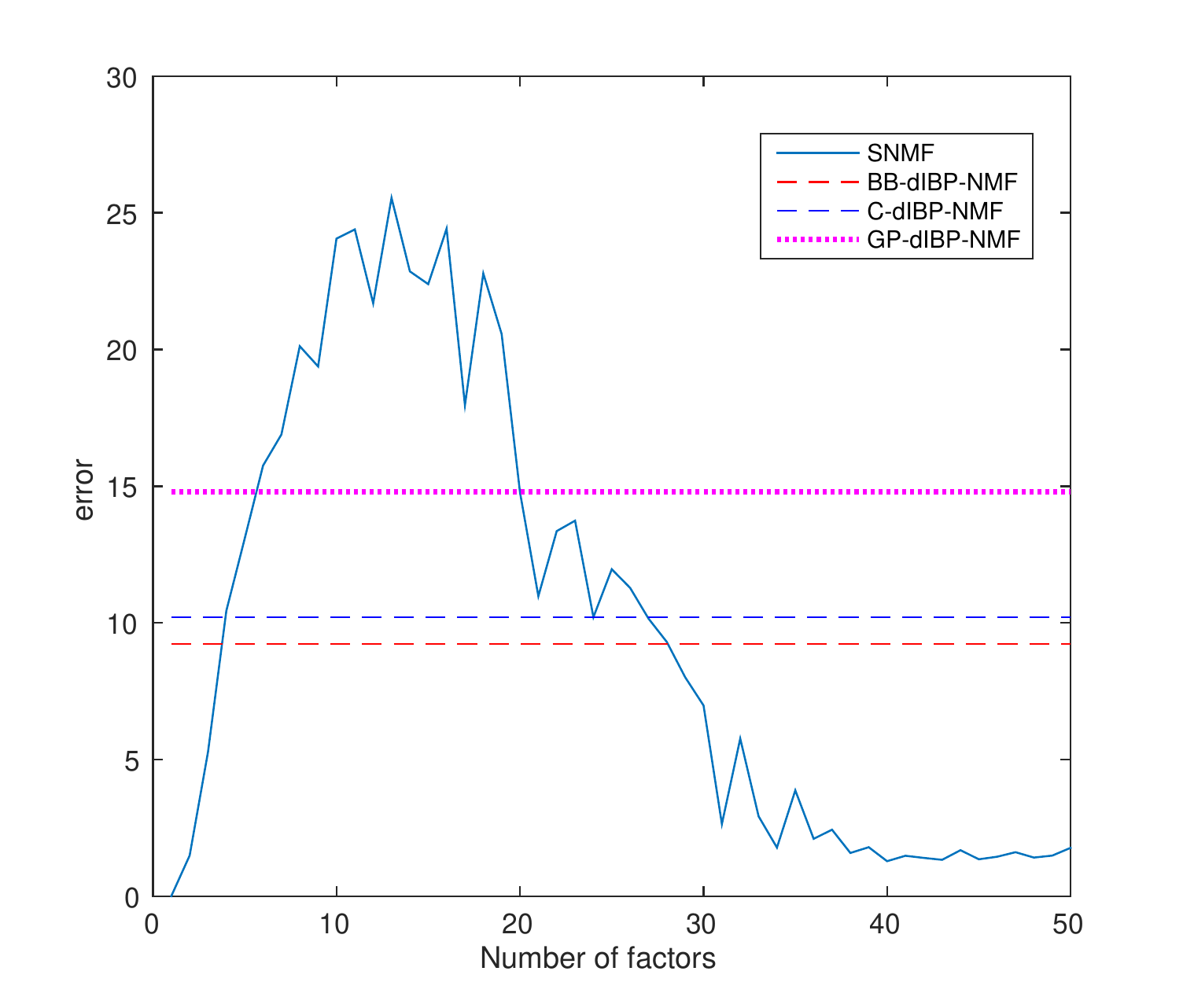}}
  \caption{Reconstruction error comparisons on synthetic dataset between GP-dIBP-NMF, BB-dIBP-NMF, C-dIBP-NMF, and Sparse NMF.}
  \label{syn-erro}
\end{figure}

\begin{figure}[t]
  \centerline{\includegraphics[scale=0.45]{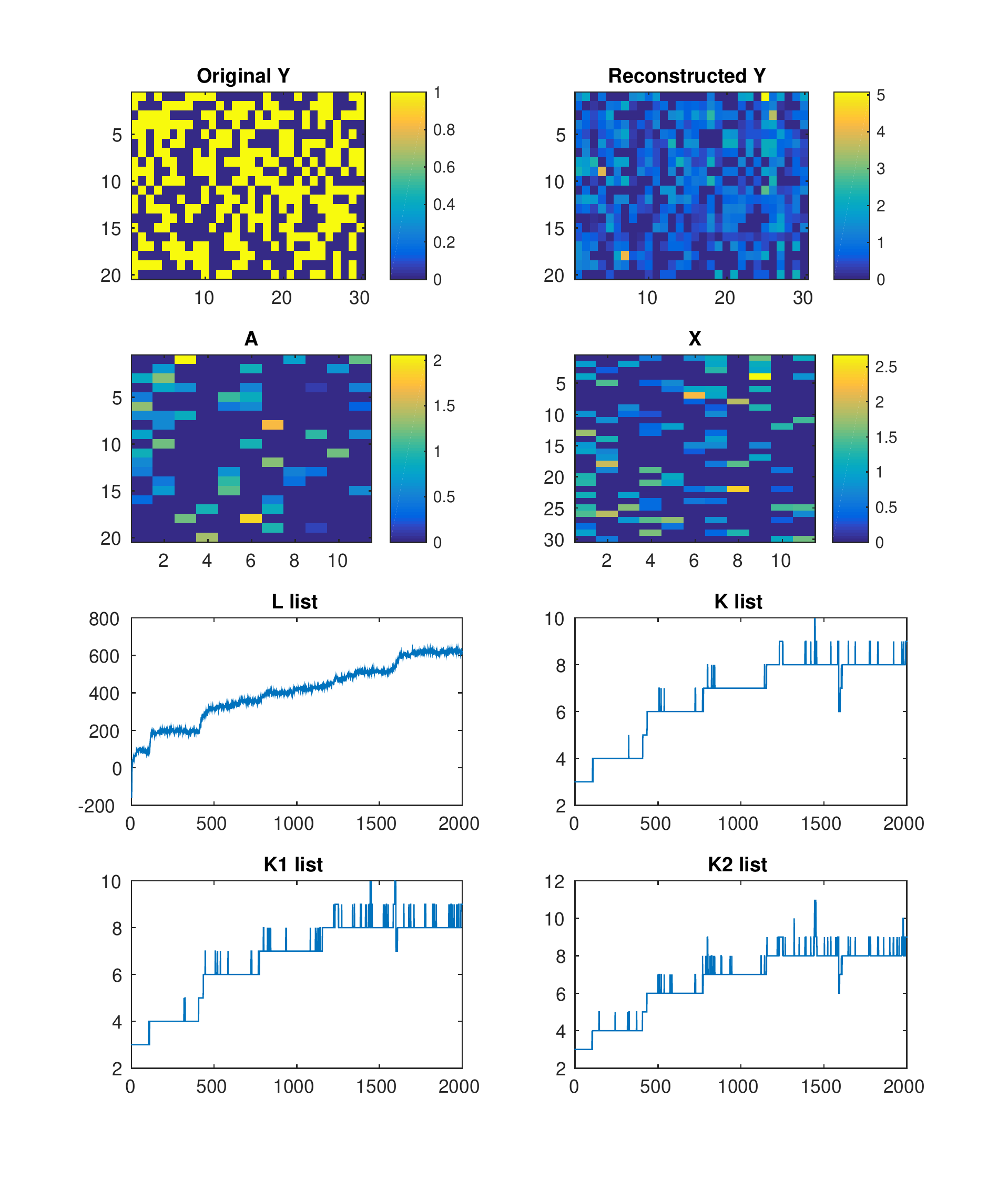}}
  \caption{Results on synthetic dataset from GP-dIBP-NMF. }
  \label{syn-gp}
\end{figure}

\begin{figure}[t]
  \centerline{\includegraphics[scale=0.4]{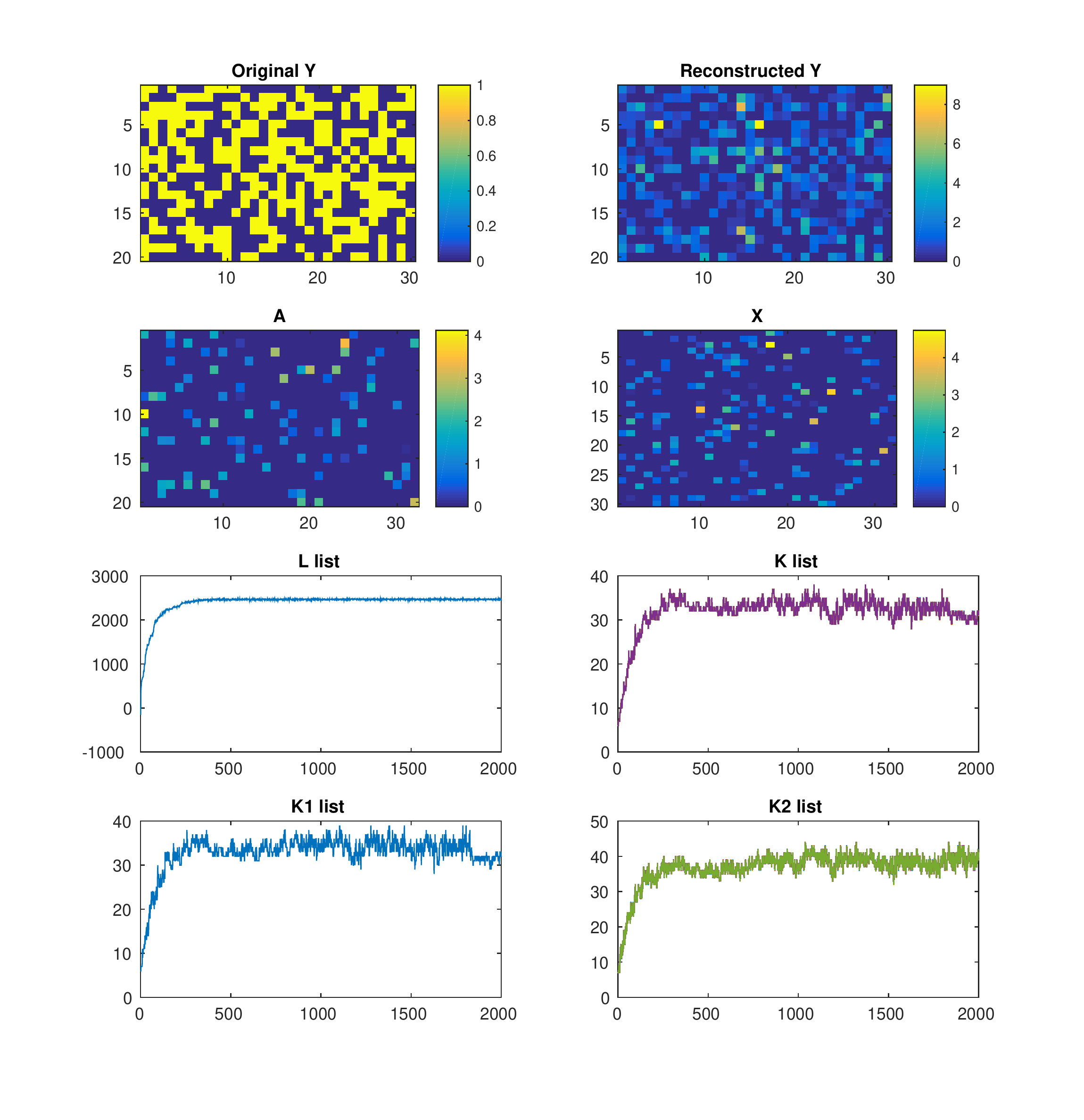}}
  \caption{Results on synthetic dataset from BB-dIBP-NMF. }
  \label{syn-bb}
\end{figure}

\begin{figure}[t]
  \centerline{\includegraphics[scale=0.45]{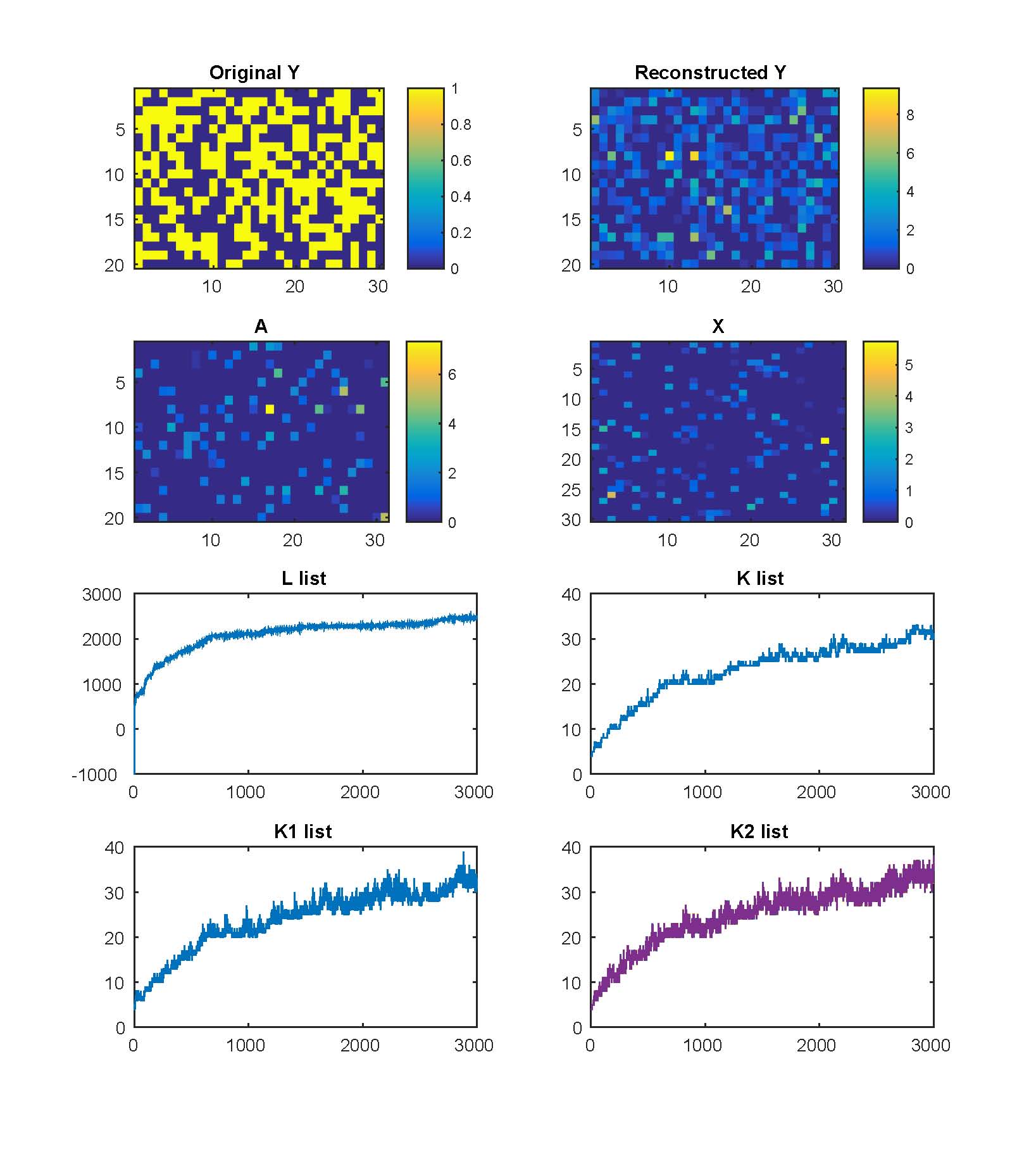}}
  \caption{Results on synthetic dataset from C-dIBP-NMF. }
  \label{syn-c}
\end{figure}

We randomly generate a matrix $Y_{20 \times 30}$ with elements in $\{0, 1\}$. The proposed algorithms: GP-dIBP-NMF, BB-dIBP-NMF and C-dIBP-NMF are used for sparse nonnegative matrix factorization on this matrix. The results are shown in Fig. \ref{syn-gp}, Fig. \ref{syn-bb} and Fig. \ref{syn-c}, respectively. In each figure, we have visualized the generated matrix and its constructed version through the learned matrices $A$ and $X$ which are both visualized in the figure. It can be observed that the (dark) positions with zero values are reconstructed well by all three algorithms comparing when the original $Y$ and the reconstructed $Y$ are compared. Another observation is the sparsity of the factor matrices $A$ and $X$ whose number of columns will change with the iteration. The last but not least observation is the value log-likelihoods from the three algorithms. A bigger log-likelihood value means that the model fits the data better. The comparison shows that the BB-dIBP-NMF has similar performance to C-dIBP-NMF but is better than GP-dIBP-NMF. Except for log-likelihood, the convergence rate of GP-dIBP-NMF is the worst of the three models, as can be observed from the convergence curve in \emph{L list} in Fig. \ref{syn-gp}, Fig. \ref{syn-bb} and Fig. \ref{syn-c}. It takes about 200 for BB-dIBP-NMF to converge, and 500 for C-dIBP-NMF to converge, but GP-dIBP-NMF needs more than 1500 iterations to reach convergence. This is due to the complex model structure of the GP-based dIBP.
As well as the visualization, the reconstruction error is also quantitatively compared. The reconstruction error is defined as
\begin{equation}
\begin{split}
error = \|Y - A * X^T\|_1
\end{split}
\end{equation}
The comparison is shown in Fig. \ref{syn-erro}. The solid (light blue) line in the figure shows the sparse NMF with different numbers of factors as the input. As shown in the figure, the worst error value is around 25. Since the proposed algorithms do not need to predefine the factor number, there are three lines in Fig. \ref{syn-erro} to indicate the errors of the three algorithms.

\begin{figure}[t]
  \centerline{\includegraphics[scale=0.6]{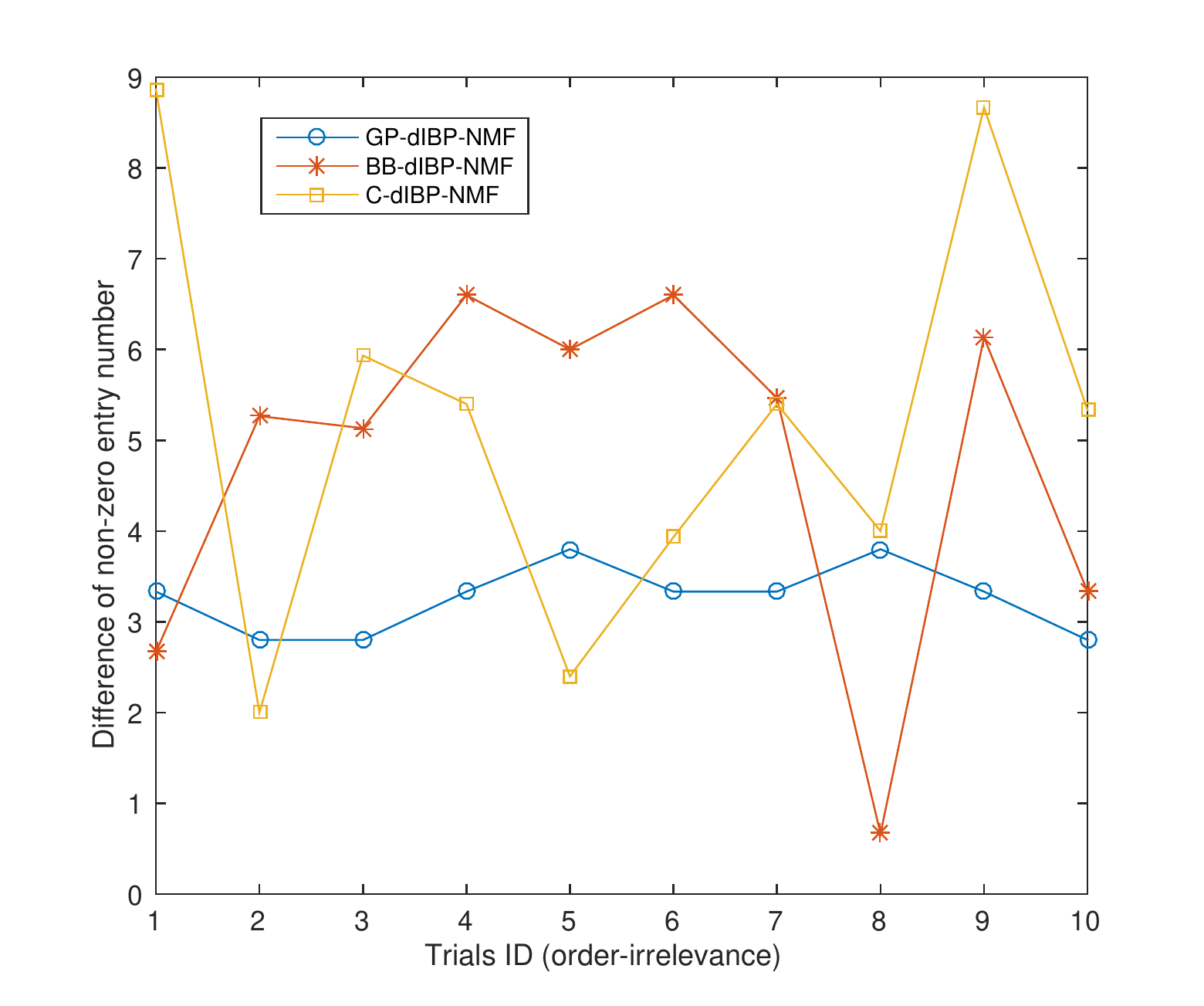}}
  \caption{Results on synthetic dataset to show the flexibility of the different models. The x-axis denotes the trial IDs (order is irrelevant) }
  \label{syn-non}
\end{figure}

In order to show the flexibility of our proposed dIBP comparing with GP-based dIBP, we firstly have designed a metric to measure this flexibility by comparing the number of non-zero entries of two loading matrices ($A$ and $X$) from the models. As we claimed, our proposed model is with more flexibility to allow the numbers of non-zero entries of loading matrices more different from each other comparing GP-based dIBP. Therefore, we have evaluated the mean of the differences between the corresponding columns of $A$ and $X$ as $\frac{1}{K}\sum_{k=1}^K |n^{(1)}_k - n^{(2)}_k|$ where $K$ is the number of columns of both matrices, $n^{(1)}_k$ is the number of non-zero entries of $k$-th column of $A$, and $n^{(2)}_k$ is the number of non-zero entries of $k$-th column of $X$. We have randomly generated ten matrices with same size: $20 \times 30$, and run three models on ten matrices. The designed metric has been evaluated by on the outputs of models. As shown in Fig. \ref{syn-non}, we can see that the BB-dIBP-NMF and C-dIBP-NMF are with more fluctuations and larger non-zero entry number differences comparing with GP-dIBP-NMF. It demonstrates that the proposed dIBPs are more flexible than GP-based dIBP.

\subsection{Real-world Dataset}

In this subsection, we apply the proposed algorithms on two real-world tasks: document clustering and recommender system. For each task, we will compare the proposed algorithms with sparse NMF with different numbers of factors.

\subsubsection{Document Clustering}

The real-world datasets used for the document clustering task are:
\begin{itemize}
  \item \textbf{Cora Dataset}\footnote{http://linqs.cs.umd.edu/projects/projects/lbc/} The Cora dataset consists of 2708 scientific publications classified into one of seven classes. Each publication in the dataset is described by a 0/1-valued word vector indicating the absence/presence of the corresponding word from the dictionary. The dictionary consists of 1433 unique words.
  \item \textbf{Citeseer Dataset} The CiteSeer dataset consists of 3312 scientific publications. Each publication in the dataset is also described by a 0/1-valued word vector indicating the absence/presence of the corresponding word from the dictionary. The dictionary consists of 3703 unique words. The labels of these papers are set as their research areas, such as AI (Artificial Intelligence), ML (Machine Learning), Agents, DB (Database), IR (Information Retrieval) and HCI (Human-Computer Interaction).
\end{itemize}

The evaluation metrics of document clustering are Jaccard Coefficient (JC), Folkes\&Mallows (FM) and F1 measure (F1). Given a clustering result,
\begin{itemize}
\item $a$ is the number of two points that are in the same cluster of both benchmark results and clustering results;
\item $b$ is the number of two points that are in the same cluster of benchmark results but in different clusters of clustering results;
\item $c$ is the number of two points that are not in the same cluster of the two benchmark results but are in the same cluster of clustering results.
\end{itemize}
and three metrics are computed the equations in Table \ref{tb:em} (bigger means better).

\begin{table}[thb]%
\centering
\renewcommand{\arraystretch}{2}
\caption{Evaluation metrics of document clustering}{%
\begin{tabular}{c|c}
\hline
Jaccard Coefficient   &~~~~$JC=\frac{a}{a+b+c}$~~~~    \\\hline
Folkes \& Mallows    &~~~~$FM=\left( \frac{a}{a+b} \cdot \frac{a}{a+c} \right)^{1/2}$~~~~    \\\hline
F1 measure     &~~~~$F1=\frac{2a^2}{2a^2+ac+ab}$~~~~        \\\hline
\end{tabular}}
\label{tb:em}
\end{table}%

\begin{figure*}[!th]
  \centerline{\includegraphics[scale=0.5]{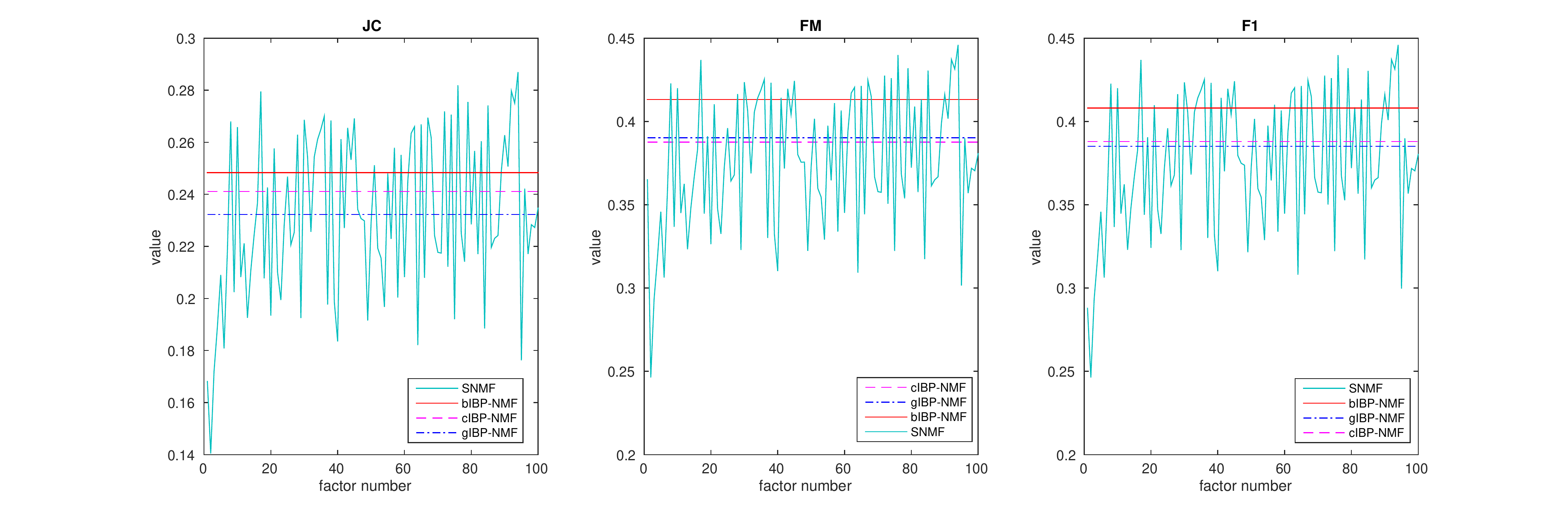}}
  \caption{Document clustering comparisons on \emph{Citeseer} dataset between GP-dIBP-NMF (gIBP-NMF), BB-dIBP-NMF (bIBP-NMF), C-dIBP-NMF (cIBP-NMF), and Sparse NMF (SNMF).}
  \label{citeseer-dc1}
\end{figure*}

\begin{figure*}[!th]
  \centerline{\includegraphics[scale=0.5]{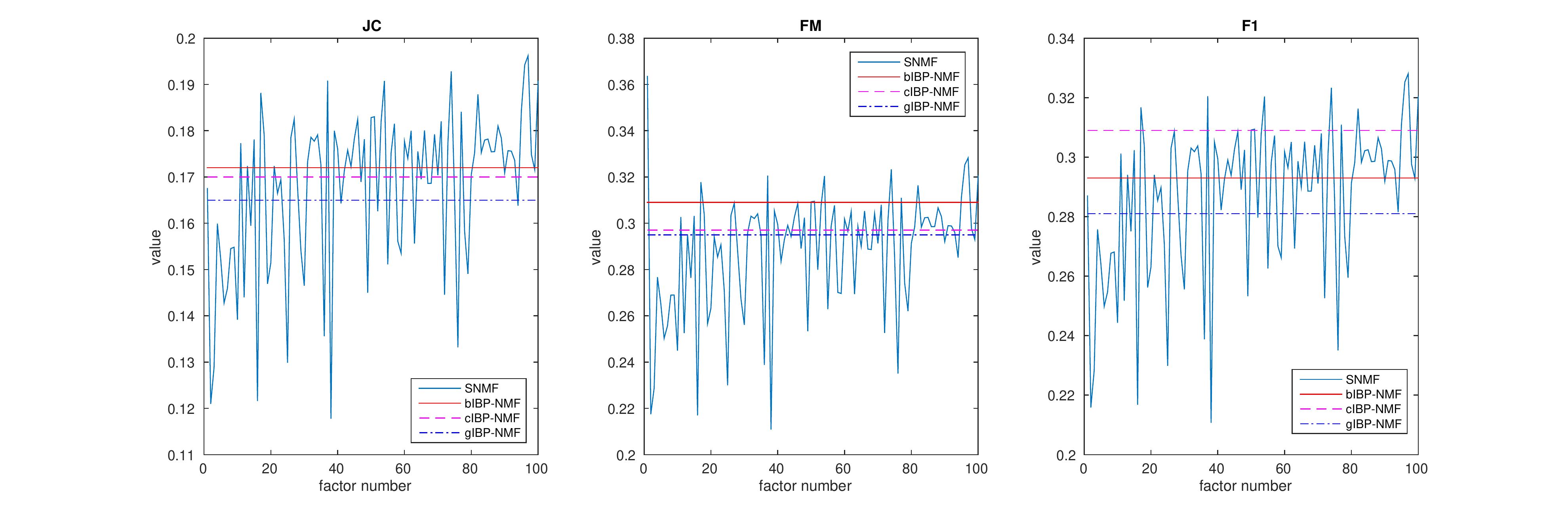}}
  \caption{Document clustering comparisons on \emph{Cora} dataset between GP-dIBP-NMF (gIBP-NMF), BB-dIBP-NMF (bIBP-NMF), C-dIBP-NMF (cIBP-NMF), and Sparse NMF (SNMF).}
  \label{cora-dc2}
\end{figure*}

When document clustering is selected as the task, the output of the algorithms ($A$ is the document-factor matrix) is used as the input of the spectral clustering algorithm. Since the common clustering algorithm is used, the clustering performance of different NMF algorithms is only determined by the factors $A$. The clustering results evaluated by the metrics in Table. \ref{tb:em} on two different datasets are shown in Fig. \ref{citeseer-dc1} and Fig. \ref{cora-dc2}. In each figure, we run the sparse nonnegative matrix factorization in Eq. \ref{nmfcf} with different factor numbers (the x-axis). All the algorithms have 1000 iterations. The results of the proposed three algorithms are also shown in the figure through three horizontal lines. The learned factor numbers are 10 (from BB-dIBP-NMF), 12 (from C-dIBP-NMF) and 12 (from GP-dIBP-NMF) for the \emph{Citeseer} dataset; 17 (from BB-dIBP-NMF), 15 (from C-dIBP-NMF) and 22 (from GP-dIBP-NMF) for the \emph{Cora} dataset. The reason why the horizontal lines are used is that the proposed algorithms do not need the factor number as an input, so the clustering results are not dependent of the x-axis (factor number). To summarize, without the prior knowledge of the number of factors, the proposed algorithms achieve relatively good performance on the document clustering task. Of the three algorithms, BB-dIBP-NMF achieves the best performance. C-dIBP-NMF and GP-dIBP-NMF achieve similar performances on two datasets only except the F1 on the \emph{Cora} dataset.

\subsubsection{Recommender System}

The dataset for this task is \textbf{MovieLens 100K\footnote{http://grouplens.org/datasets/movielens/}}. This data set consists of 100,000 ratings (1-5) from 943 users on 1682 movies, and each user has rated at least 20 movies. In the following experiment, we use a 5-fold method for cross validation ($20\%$ ratings are kept as the test data and the remaining $80\%$ ratings are used as the training data).

When used for the recommender system, the outputs of the algorithms $A$ (user-factor matrix) and $X$ (movie-factor matrix) reconstruct the rating matrix $Y_r=A * X^T$ on the retained test ratings. The quantitative evaluation is
\begin{equation}
MAE = \|Y_r - Y_{test}\|_1
\end{equation}
Here, we do not normalize $MAE$ since it does not impact on the comparison. Note that the sparse NMF in Eq. \ref{nmfcf} needs to be revised to ignore the test ratings. The weighted NMF \cite{kim2009weighted,guillamet2003introducing} can be adopted and a mask matrix with ones on the training ratings and zeros on the test ratings is constructed. Ignoring the test ratings can be easily achieved by only considering the training ratings in the updates for $Z$ in Eq. \ref{bbnmfz1} and Eq. \ref{bbnmfz2} and updates for $V$ in Eq. \ref{nmflikelihoodv1} and Eq. \ref{nmflikelihoodv2}.

\begin{figure*}[!th]
  \centerline{\includegraphics[scale=0.52]{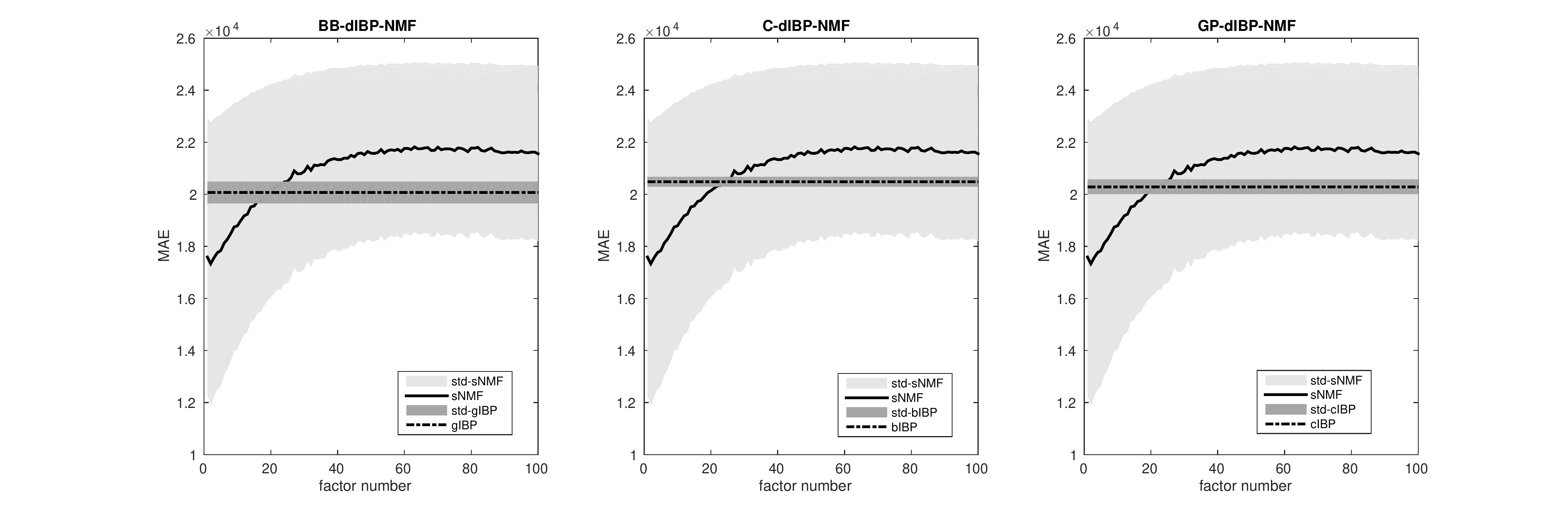}}
  \caption{Recommendation comparisons on \emph{MovieLens-100K} dataset between GP-dIBP-NMF (gIBP-NMF), BB-dIBP-NMF (bIBP-NMF), C-dIBP-NMF (cIBP-NMF), and Sparse NMF (SNMF).}
  \label{ml100k-rs1}
\end{figure*}

The recommendation results are shown in Fig. \ref{ml100k-rs1}. Considering the 5-fold cross test, the mean of the five groups is plotted in the figure with a solid line, and the standard deviations are shown in the figure by the gray area around the mean plot. Three subfigures in Fig. \ref{ml100k-rs1} denote the comparison of the sparse NMF with the proposed three algorithms, for which the results are plotted by dashed lines. Similarly, the standard deviations are also given by the area around the dashed lines. It can be observed that the SNMF has better performance than the proposed algorithms only when the factor number is smaller than 20. The proposed algorithms outweigh the SNMF (smaller MAE values) for the remaining choices for the factor number. We want to highlight again that the factor number is unknown before running SNMF. There is a wide possible range for this number. Within this wide range, there are only approximately 20 choices that give better performance than the proposed algorithms. Three algorithms have similar performance to one another. The BB-dIBP-NMF algorithm is the best, with relatively smaller MAE, and C-dIBP-NMF is the worst. The C-dIBP-NMF, however, has the smallest variance. The small variance means that this algorithm achieves similar performance on different test data. In other words, an algorithm with small variance will be stable on different test data. We can see that SNMF is not stable compared with the proposed algorithms.


\section{Conclusion and Further Study}

The renowned nonnegative matrix factorization is advantageous for many machine learning tasks, but the assumption that the dimension of the factors is known in advance makes NMF impractical for many applications. To resolve this issue, we have proposed a nonparametric NMF framework based on dIBP to remove the assumption. First, a model is built by implementing this framework using GP-based dIBP, which successfully removes the assumption but suffers from larger model complexity and less flexibility. Then, we have proposed two new dIBPs through a bi-variate beta distribution and a copula. One advantage of the models based on new dIBPs is that they have simpler model structures than models with GP-based dIBP. Lastly, three inference algorithms have been designed for the proposed three models, respectively. The experiments on the synthetic and real-world datasets demonstrates the capability of the proposed models to perform NMF without predefining the dimension number, and the models based on the new dIBPs have better convergence rates and more flexibility than the model based on GP-based dIBP.

The bivariate beta distribution-based model and copula-based model have achieved comparative performances both on document clustering and recommender system. We here give hints for their usage: 1) if the data correlation is within the range $[-0.3, 0.3]$, the FGM copula-based model will have better flexibility than the bivariate beta distribution-based model; 2) if the data correlation is outside the range of $[-0.3, 0.3]$, the bivariate beta distribution-based model is more reasonable than the FGM copula-based model.

One possible future study for this work is the aspect of efficiency. Current Gibbs sampling inference is not efficient enough for big data. Our further study will focus on the efficiency of the inference of the proposed models using the variational inference strategy.

\section*{Acknowledgment}

The research work reported in this paper was partly supported by the Australian Research Council (ARC) under discovery grant DP140101366 and the China Scholarship Council. This work was jointly supported by the National Science Foundation of China under grant no.61471232.


\bibliographystyle{IEEEtran}
\bibliography{TNNLS}

\appendices

\section{The conditional distributions for the GP-based dIBP NMF}

The conditional distributions are:

\textbf{Sampling} $\mu$
\begin{equation}
\begin{aligned}
p(\mu_k | \cdots) \propto \frac{\mu^{\alpha}_K}{\mu_k} \prod_{t=1}^{2}\prod_n^{N_t}
(\gamma^t_k)^{z^t_{n,k}} (1 - \gamma^t_k)^{1-z^t_{n,k}}
\end{aligned}
\label{gpnmfmu}
\end{equation}
where
\begin{equation}
\begin{aligned}
\gamma^t_k = F(F^{-1}(\mu_k | 0, \Sigma_k^{(t,t)} + \eta^2  ) - g^t_k | 0, \eta^2 )
\end{aligned}
\label{gpnmfgamma}
\end{equation}
where $F()$ is normal cumulative distribution function.

\textbf{Sampling} $g$

\begin{equation}
\begin{aligned}
p(g_k | \cdots) \propto \mathcal{N}(g_k | 0, \Sigma_k) \cdot \prod_t \prod_n^{N_t} \mathcal{N}(h^t_{n,k} | g^t_k, \eta^2)
\end{aligned}
\label{gpnmfg}
\end{equation}
where $\mathcal{N}()$ denotes normal distribution.

\textbf{Sampling} $h$
\begin{equation}
\begin{aligned}
p(h^t_{n,k} | \cdots) \propto \mathcal{N}(g^t_k, \eta^2)
\end{aligned}
\label{gpnmfh}
\end{equation}
with support
\begin{equation}
\begin{aligned}
\left\{
  \begin{array}{l l}
    h^t_{n,k} \in (-\infty, \widetilde{\mu}_k^t] & \quad \text{if $z^t_{n,k}=1$}
    \\
    h^t_{n,k} \in [\widetilde{\mu}_k^t, +\infty) & \quad \text{if $z^t_{n,k}=0$}
  \end{array} \right.
\end{aligned}
\end{equation}
where
\begin{equation}
\begin{aligned}
\widetilde{\mu}_k^t = F^{-1}(\mu_k | 0, \Sigma_k^{(t, t)} + \eta^2)
\end{aligned}
\end{equation}

\textbf{Sampling} $s$
\begin{equation}
P(s | \cdots) \propto gamma(s; hs, 1) \prod_k^K \mathcal{N}(g_k | 0, \Sigma_k)
\label{gpnmfs}
\end{equation}

\textbf{Sampling} $Z$
\begin{equation}
\begin{aligned}
&p(z_{m,k}^{(1)} = 1 | \cdots) \propto \gamma^1_k
\\
& \prod_n Exp(y_{m, n}; \sum_{l} v^{(1)}_{m, l}  z^{(1)}_{m, l}  v^{(2)}_{n, l}  z^{(2)}_{n, l} + \epsilon)
\\
&p(z_{m,k}^{(1)} = 0 | \cdots) \propto (1 - \gamma^1_k)
\\
&\prod_n Exp(y_{m, n}; \sum_{l} v^{(1)}_{m, l}  z^{(1)}_{m, l}  v^{(2)}_{n, l}  z^{(2)}_{n, l} + \epsilon)
\end{aligned}
\label{gpnmfz1}
\end{equation}
and
\begin{equation}
\begin{aligned}
&~~~p(z_{n,k}^{(2)} = 1 | \cdots)
\\
&\propto \gamma^2_k \prod_m Exp(y_{m, n}; \sum_{l} v^{(1)}_{m, l}  z^{(1)}_{m, l}  v^{(2)}_{n, l}  z^{(2)}_{n, l} + \epsilon)
\\
&~~~p(z_{n,k}^{(2)} = 0 | \cdots)
\\
&\propto (1 - \gamma^2_k) \prod_m Exp(y_{m, n}; \sum_{l} v^{(1)}_{m, l}  z^{(1)}_{m, l}  v^{(2)}_{n, l}  z^{(2)}_{n, l} + \epsilon)
\end{aligned}
\label{gpnmfz2}
\end{equation}
where $\gamma^t_k$ is same as in Eq. (\ref{gpnmfgamma}).

\begin{IEEEbiography}[{\includegraphics[width=1in,height=1.25in,clip,keepaspectratio]{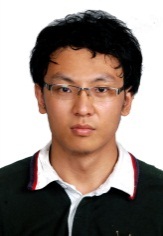}}]{Junyu Xuan}
received the bachelor's degree in 2008 from China University of Geosciences, Beijing. Currently he is working toward the dual-doctoral degree in both Shanghai University and the University of Technology, Sydney. His main research interests include Machine Learning, Complex Network and Web Mining.
\end{IEEEbiography}

\begin{IEEEbiography}[{\includegraphics[width=1in,height=1.25in,clip,keepaspectratio]{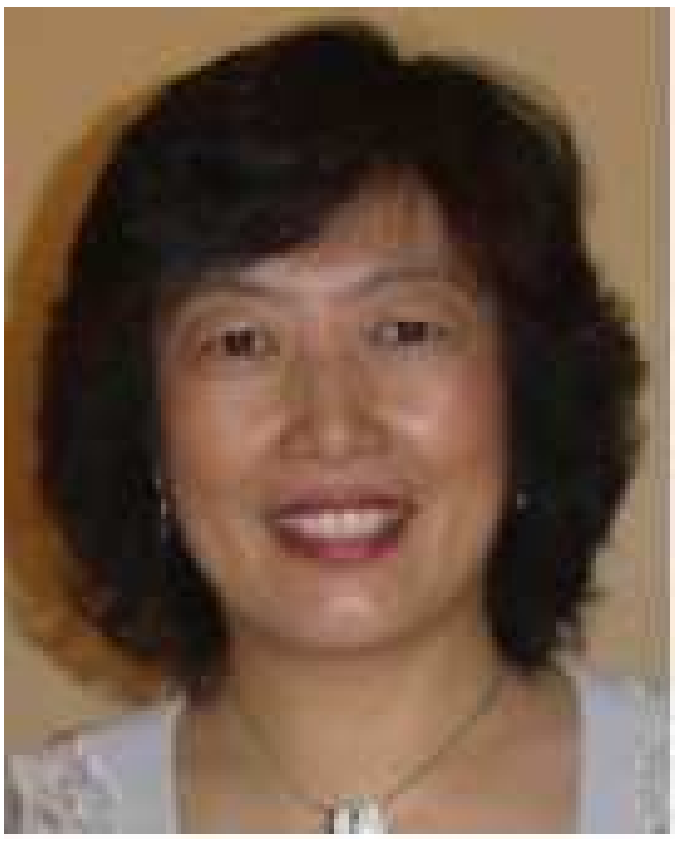}}]{Jie Lu}
is a full professor and Associate Dean in the Faculty of Engineering and Information Technology (FEIT) at the University of Technology, Sydney. Her research interests lie in the area of decision support systems and uncertain information processing. She has published five research books and 350 papers, won 7 Australian Research Council discovery grants and 10 other grants. She received a University Research Excellent Medal in 2010. She serves as Editor-In-Chief for Knowledge-Based Systems (Elsevier), Editor-In-Chief for International Journal of Computational Intelligence Systems (Atlantis), editor for book series on Intelligent Information Systems (World Scientific) and guest editor of six special issues for international journals, as well as delivered six keynote speeches at international conferences.
\end{IEEEbiography}

\begin{IEEEbiography}[{\includegraphics[width=1in,height=1.25in, clip]{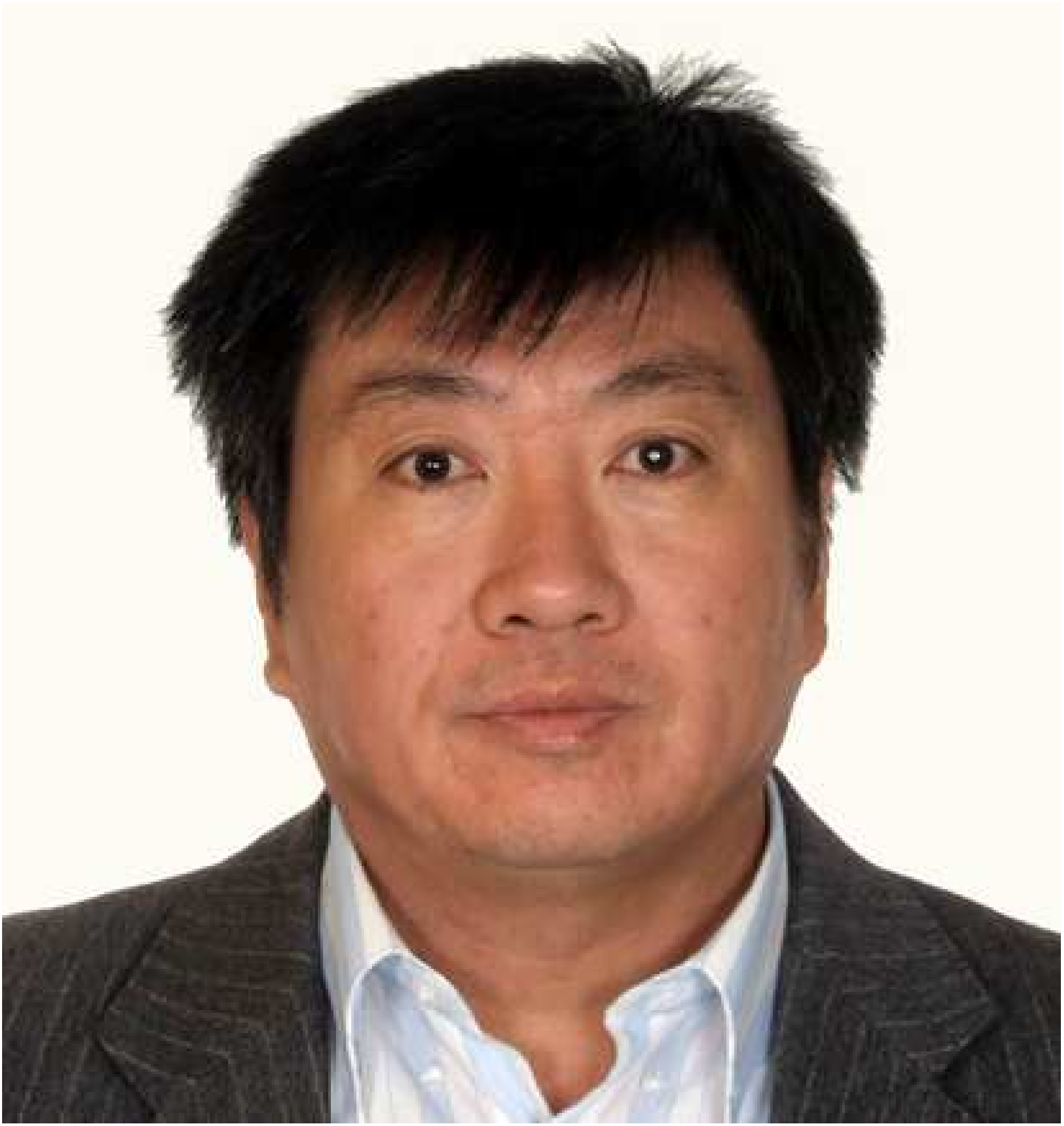}}]{Guangquan Zhang}
is an associate professor in Faculty of Engineering and Information Technology at the University of Technology Sydney (UTS), Australia. He has a PhD in Applied Mathematics from Curtin University of Technology, Australia. He was with the Department of Mathematics, Hebei University, China, from 1979 to 1997, as a Lecturer, Associate Professor and Professor. His main research interests lie in the area of multi-objective, bilevel and group decision making, decision support system tools, fuzzy measure, fuzzy optimization and uncertain information processing. He has published four monographs, four reference books and over 350 papers in refereed journals and conference proceedings. He has won 6 Australian Research Council (ARC) discovery grants and many other research grants.
\end{IEEEbiography}

\begin{IEEEbiography}[{\includegraphics[width=1in,height=1.25in, clip]{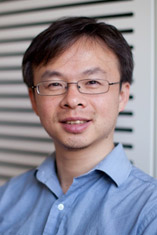}}]{Richard Yi Da Xu}
received the B.Eng. degree in computer engineering from the University of New
South Wales, Sydney, NSW, Australia, in 2001, and the Ph.D. degree in computer sciences from the University of Technology at Sydney (UTS), Sydney, NSW, Australia, in 2006. He is currently a Senior Lecturer with the School of Computing and Communications, UTS. His current research interests include machine learning, computer vision, and statistical data mining.
\end{IEEEbiography}

\begin{IEEEbiography}[{\includegraphics[width=1in,height=1.25in,clip,keepaspectratio]{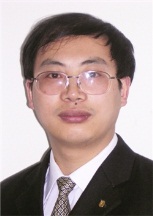}}]{Xiangfeng Luo}
is a professor in the School of Computers, Shanghai University, China. Currently, he is a visiting professor in Purdue University. He received the master's and PhD degrees from the Hefei University of Technology in 2000 and 2003, respectively. He was a postdoctoral researcher with the China Knowledge Grid Research Group, Institute of Computing Technology (ICT), Chinese Academy of Sciences (CAS), from 2003 to 2005. His main research interests include Web Wisdom, Cognitive Informatics, and Text Understanding. He has authored or co-authored more than 50 publications and his publications have appeared in IEEE Trans. on Automation Science and Engineering, IEEE Trans. on Systems, Man, and Cybernetics-Part C, IEEE Trans. on Learning Technology, Concurrency and Computation: Practice and Experience, and New Generation Computing, etc. He has served as the Guest Editor of ACM Transactions on Intelligent Systems and Technology. Dr. Luo has also served on the committees of a number of conferences/workshops, including Program Co-chair of ICWL 2010 (Shanghai), WISM 2012 (Chengdu), CTUW2011 (Sydney) and more than 40 PC members of conferences and workshops.
\end{IEEEbiography}

\end{document}